\documentclass{article}

\usepackage[main, final]{neurips_2026}

\usepackage[utf8]{inputenc} 
\usepackage[T1]{fontenc}    
\usepackage{url}            
\usepackage{booktabs}       
\usepackage{amsfonts}       
\usepackage{nicefrac}       
\usepackage{microtype}      
\usepackage{xcolor}         
\usepackage{amsthm}
\usepackage{amssymb}
\usepackage{graphicx}
\usepackage{caption}
\usepackage{amsmath}
\usepackage[utf8]{inputenc}
\usepackage[T1]{fontenc}
\usepackage{url}
\usepackage{booktabs}
\usepackage{amsfonts}
\usepackage{textalpha}
\usepackage{nicefrac}
\usepackage{microtype}
\usepackage{xcolor}
\usepackage{graphicx}
\usepackage{caption}
\usepackage{subcaption}
\usepackage{multirow}
\usepackage{makecell}
\usepackage{pifont}
\usepackage{enumitem}
\usepackage{amsmath}
\usepackage{amssymb}
\usepackage{wrapfig}
\usepackage{xspace}
\usepackage{arydshln}
\usepackage[table]{xcolor}
\usepackage{colortbl} 
\usepackage{graphicx}
\usepackage{arydshln}
\setlist[enumerate]{leftmargin=*, label=(\arabic*), topsep=0pt}
\setlist[itemize]{leftmargin=*, topsep=0pt}
\usepackage{tikz}
\definecolor{yellowtext}{RGB}{68,132,243}
\definecolor{yellowred}{RGB}{50,167,82}
\definecolor{yellowblue}{RGB}{251,191,5}

\newcommand{\stddev}[1]{\textcolor{gray}{\scalebox{.8}{$\pm$#1}}}
\definecolor{darkblue}{rgb}{0, 0, 0.5}
\definecolor{chocolate}{HTML}{D2691E}
\definecolor{maroon}{HTML}{A00000}
\definecolor{indigo}{HTML}{4B0082}
\definecolor{violet}{HTML}{4B2E83}
\definecolor{lightblue}{rgb}{0.0, 0.0, 0.5}
\definecolor{cadmiumgreen}{rgb}{0.0, 0.42, 0.24}
\definecolor{forestgreen}{rgb}{0.13, 0.55, 0.13}
\usepackage[most]{tcolorbox}
\usepackage{fvextra}
\DefineVerbatimEnvironment{VerbatimWrap}{Verbatim}{
breaklines=true, 
breakindent=0pt, 
breaksymbol={},
fontsize=\scriptsize,
}
\newcommand{\eg}{e.g.,\xspace}
\newcommand{\ie}{i.e.,\xspace}
\newcommand{\method}{\textit{\textbf{SMART-Med}}\xspace}
\newcommand{\dataset}{\textit{\textbf{MedVR-3K}}\xspace}
\usepackage[framemethod=tikz]{mdframed}
\mdfdefinestyle{customstyle}{
linecolor=cyan!70,
linewidth=3pt,
innerleftmargin=5pt,
topline=false,
rightline=false,
bottomline=false,
leftline=true,
innerrightmargin=5pt,
innertopmargin=5pt,
innerbottommargin=5pt,
backgroundcolor=cyan!8,
}
\newenvironment{custommdframed}
{\begin{mdframed}[style=customstyle]\footnotesize}
{\end{mdframed}}
  
\usepackage{longtable}
\usepackage{tabularx}
\usepackage{array}
\newcolumntype{P}[1]{>{\raggedright\arraybackslash}m{#1}}
\usepackage{colortbl}

\definecolor{lightgray}{rgb}{0.9, 0.9, 0.9}

\usepackage{tikz}

\usepackage[colorlinks=true,linkcolor=blue,citecolor=green!50!black,urlcolor=magenta!70!black,filecolor=cyan!70!black]{hyperref}
\usepackage{graphicx}
\usepackage{listings}
\usepackage{pifont}  
\usepackage{underscore}
\usepackage{wrapfig}
\newcommand{\cmark}{\ding{51}}  
\newcommand{\xmark}{\ding{55}}  

\lstdefinelanguage{json}{
  basicstyle=\ttfamily\small,
  numbers=left,
  numberstyle=\tiny,
  stepnumber=1,
  numbersep=6pt,
  showstringspaces=false,
  breaklines=true,
  frame=single,
  rulecolor=\color{black!20},
  string=[s]{"}{"},
  stringstyle=\color{black!70},
}

\title{Learning When to Collaborate: Selective Multi-Agent Medical Reasoning via Uncertainty-Aware Routing}

\author{%
  Jiacheng Hou\textsuperscript{1}, Xueliang Cui\textsuperscript{2,4}, Dan Lu\textsuperscript{5}, Ruxin Wang\textsuperscript{3,4}\thanks{Corresponding author. Mails to rx.wang@siat.ac.cn and houjc24@mails.tsinghua.edu.cn.} \\[4pt]
  \textsuperscript{1}SIGS, Tsinghua University \quad
  \textsuperscript{2}Southern University of Science and Technology \\
  \textsuperscript{3}Shenzhen Institutes of Advanced Technology, Chinese Academy of Sciences \\
  \textsuperscript{4}Shenzhen University of Advanced Technology, China\\
  \textsuperscript{5}West China Hospital, Sichuan University, China. \\[4pt]
  \url{https://github.com/JayceonHo/SMART-Med}
}

\begin{document}

\maketitle

\begin{abstract}
Multi-agent systems have emerged as a promising paradigm for multimodal medical reasoning, but existing methods rely on two assumptions that hinder practical deployment: (i) always-active dense collaboration among multiple agents, and (ii) dependence on large cloud-based models that raise privacy and efficiency concerns. In this paper, we challenge both assumptions and propose \method, a fully local multi-agent method that treats collaboration as a learnable decision rather than a default behavior. \emph{Our key insight is that not every medical query requires costly multi-agent deliberation: many can be reliably solved by a single well-trained agent, while only uncertain cases benefit from expert collaboration.} Based on this insight, \method introduces an uncertainty-driven routing mechanism that adaptively decides when to answer directly and when to recruit a small subset of expert agents, effectively converting dense multi-agent reasoning into selective, query-adaptive coordination. To make this routing reliable, we curate a new multimodal medical dataset \dataset with high-quality reasoning traces and a difficulty hierarchy, and then fine-tune local agents using GRPO with two complementary rewards: a certainty-aware accuracy reward and a vision-language reward model for correctness and reasoning quality respectively. Experiments on four medical visual question answering benchmarks show that \method achieves competitive performance compared to existing MAS approaches while substantially reducing inference cost (reducing inference time by more than 3.8$\times$ and token usage by 72\%). Moreover, after fine-tuning on \dataset, \method yields an average improvement of 6.7\% across Slake, PathVQA, VQA-RAD, and PMC-VQA. Our results suggest that selective collaboration can be an efficient yet effective alternative to dense collaboration and enable practical, privacy-preserving deployment of multi-agent medical AI. 
\end{abstract}

\section{Introduction}
Multi-Agent Systems (MAS) have recently emerged as a promising paradigm for complex reasoning in various tasks~\cite{chanchateval,chen2023multi}. In such systems, multiple agents, typically instantiated by Multimodal Large Language Models (MLLMs), are assigned specialized roles and tasks, interacting through structured communication protocols to collaboratively solve complex tasks. Recently, this paradigm has demonstrated promising performance in Medical Visual Question Answering (MedVQA) and broader multimodal medical reasoning~\cite{kim2024mdagents, tang2024medagents, Xia2026MMedAgentRL}. Despite these advances, two critical limitations hinder the practical deployment of MAS in real-world clinical settings.

\begin{figure}[!htb]
    \centering
    \includegraphics[width=\linewidth]{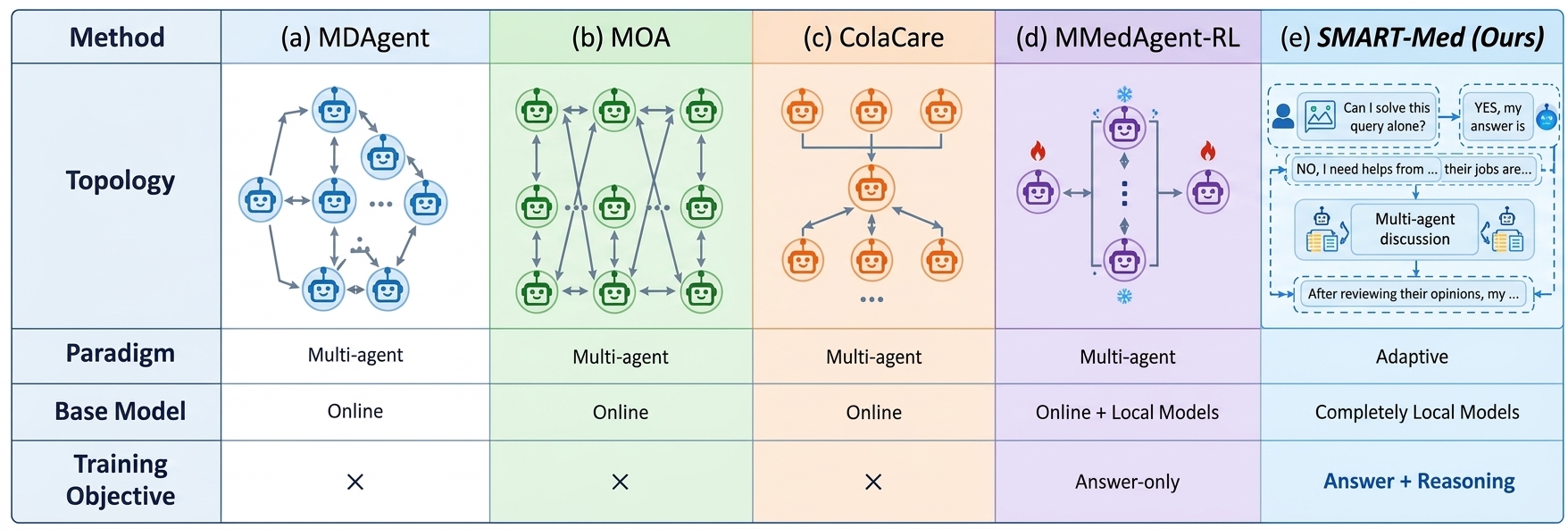}
    \caption{Comparison between existing MAS and our proposed \method. (a)-(e) respectively depict the collaboration topology, paradigm, base model and training objective of different approaches.}
    \label{fig:teaser}
\end{figure}

First, existing MAS predominantly rely on large-scale, cloud-based MLLMs (\eg GPT~\cite{gpt3inhealthcare}, Gemini~\cite{geminiteam2024}), raising concerns regarding data privacy, security, and regulatory compliance. Medical data, particularly multimodal patient records such as MRI, CT, and X-ray scans, are sensitive, and transmitting such data to external servers is often restricted or prohibited in clinical practice. This fundamentally limits the applicability of cloud-dependent MAS solutions. Second, as compared in Fig.~\ref{fig:teaser} (a)-(d), current MAS typically adopt dense communication topologies, where multiple agents engage in multi-round interactions for every query. While this aims to complementarily enhance reasoning quality, they introduce substantial computational overhead and inference latency, making them impractical for resource-constrained clinical environments. Recent studies~\cite{zhumedagentboard} show that dense collaboration often yields fairly marginal gains on MedQA and MedVQA benchmarks, questioning the necessity of full-agent participation for every query.

To address these challenges, we propose \method, which operates entirely by local, small-scale MLLMs and incorporates an adaptive collaboration mechanism. Our key insight is that multi-agent collaboration should be \emph{selective rather than universal}: a certain number of queries that can be reliably solved by a single agent, while only a subset of complex cases require coordinated reasoning. 

Based on this insight, \method introduces two complementary technical components. First, as depicted in Fig.~\ref{fig:teaser} (e), we develop an uncertainty-driven routing mechanism that prevents multi-agent collaboration when the primary agent is certain of its response. In particular, queries with high confidence are answered directly; only uncertain cases trigger dynamic selection of a small subset of expert agents for targeted reasoning. This design significantly reduces inference cost and latency while preserving the benefits of collaboration when necessary.

Second, all agents in \method are initialized with local small-scale MLLMs, with fine-tuning applied to improve domain-specific capability. To achieve this, we first construct \dataset based on existing MedVQA datasets, while \dataset additionally offers high-quality reasoning chains distilled from advanced MLLMs with multi-step filtering, and an independent judge MLLM is used to filter reasoning traces and assign difficulty scores, yielding a \emph{hierarchical and quality-controlled reasoning-centric dataset}. 

Building on this, we fine-tune the local agent using Group Relative Policy Optimization (GRPO) with two tailored rewards. For answer correctness, we introduce a \emph{certainty-aware accuracy reward} that explicitly couples prediction confidence with correctness, encouraging the model to be confident when correct and uncertain otherwise. For reasoning quality, we employ a vision-language reward model, trained on \dataset, to provide fine-grained feedback on the generated reasoning process. Together, these two components improve both reasoning capability and uncertainty calibration, which are critical for enabling reliable selective collaboration.  Our contributions can be summarized:

\begin{itemize}
    \item We propose \method, the first medical multi-agent system for MedVQA that operates entirely with local, small-scale MLLMs via an uncertainty-driven selective collaboration mechanism, dynamically triggering multi-agent reasoning only for uncertain cases. This design ensures data 
    privacy, significantly reduces inference cost and latency, while preserving collaborative benefits.
    
    \item We construct \dataset, a hierarchical reasoning-centric dataset. With the aid of \dataset, we fine-tune the local model using GRPO with two complementary new rewards: a certainty-aware accuracy reward that couples confidence with correctness, and a vision-language reward model that provides fine-grained reasoning quality feedback. 
    
    \item Extensive experiments on four public MedVQA benchmarks (\ie Slake, PathVQA, VQA-RAD, and PMC-VQA) show that \method achieves competitive or superior performance compared to existing multi-agent frameworks, while significantly reducing inference cost. 
\end{itemize}

\section{Related Work}

\paragraph{Medical Visual Question Answering.}
MedVQA aims to automatically answer clinically relevant questions about medical images, requiring precise integration of visual perception and domain knowledge. Conventional MedVQA methods relied on task-specific pipelines. Chen et al. combined CNN-based visual encoding with question classification and retrieval-based answer selection to address long-tailed distributions \cite{chen2020hcp}. Later on, attention-based fusion and consistency constraints are leveraged for improved cross-modal interaction \cite{huang2023dual, tascon2022consistency}. Recent advances shifted to Transformer-based architectures and vision-language pretraining, leveraging cross-modal attention, contrastive learning \cite{li2023self, lin2023pmc, zhang2023biomedclip}, and masked multimodal objectives \cite{li2023masked} to enhance representation alignment. More recently, foundation models such as LLaVA-Med \cite{li2023llavamed} and Med-Flamingo \cite{moor2023medflamingo} enable open-ended and few-shot medical reasoning via instruction tuning and generative modeling, yet still suffer from hallucination and limited clinical reliability, motivating extensions such as retrieval-augmented generation (MMed-RAG \cite{Xia2025MMedRAG}) and structured reasoning frameworks.

\paragraph{Multi-Agent Collaboration in Medical Analysis.}
To address the limitations of single-model reasoning, multi-agent systems introduce collaborative inference through role-based interaction and consensus mechanisms. Early approaches rely on predefined discussion protocols or voting strategies, improving robustness but lacking adaptability and efficiency in complex clinical scenarios~\cite{ chen2024reconcile, tang2024medagents, wang2024mixture}. Subsequent work explores more flexible coordination strategies. SMOA~\cite{li2025smoa} proposes a sparser collaboration structure by ignoring the results of some agents, producing even better performance. MDAgents \cite{kim2024mdagents} dynamically determine the collaboration structures of agents based on task complexity. Despite these advances, most existing methods still depend on heuristic interaction designs or incur substantial coordination overhead, motivating recent learning-based approaches such as MMedAgent-RL \cite{Xia2026MMedAgentRL}, which optimize agent collaboration and information fusion through reinforcement learning for more scalable and reliable medical reasoning.

\section{Methodology}

\subsection{Problem Formulation}
In MedVQA, the input consists of a medical image $x \in \mathcal{X}$ (\eg X-ray, CT, MRI, pathology slides) and a natural language question $q \in \mathcal{Q}$. The goal is to generate an answer $y \in \mathcal{Y}$ accompanied by a reasoning chain $r$. In a multi-agent system, we define a set of $K$ agents $\mathcal{A} = \{A_1, \dots, A_K\}$, where each agent is instantiated by a MLLM with parameters $\theta_k$. Agents may specialize in different reasoning styles or roles (\eg visual grounding, clinical knowledge, verification). Current MAS generally produce answers following a dense collaboration protocol:
\begin{equation}
y = \mathcal{F}_{\text{dense}}(x, q; \mathcal{A}),
\end{equation}
where multiple agents engage in multi-round communication. In contrast, we aim to learn a routing policy $\pi_\theta$ that smartly switches between single-agent and multi-agent modes:
\begin{equation}
y = \mathcal{F}_{\text{sparse}}(x, q; \mathcal{A}, \pi_\theta),
\end{equation}
where $\pi_\theta$ determines whether multi-agent collaboration is needed and, if so, how many agents should be activated, conditioned on the input. This transforms the dense communication topology of a MAS into a sparse, query‑adaptive structure by activating only a few (or even a single) agent.

\begin{figure}[t]
    \centering
    \includegraphics[width=\textwidth]{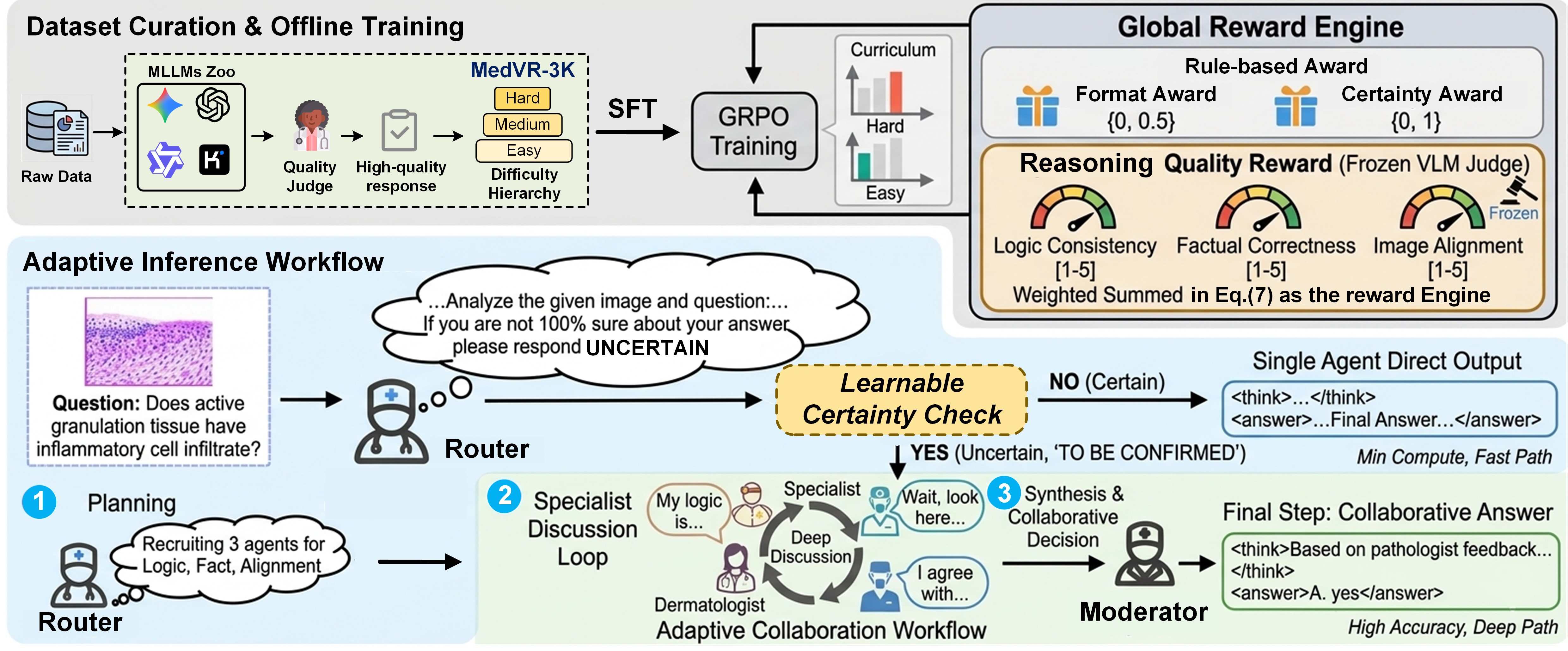}
    \caption{Overview of \method. The framework unifies reasoning-centric training and uncertainty-aware routing in a closed loop: \dataset and reward-guided optimization improve both reasoning capability and uncertainty calibration, while the learned router adaptively decides whether to answer directly or trigger selective expert collaboration.}
    \label{fig:overview}
\end{figure}

\subsection{\method: Learning When to Collaborate}
\label{sec:method}
In real-world clinical practice, the first step when a patient visits a hospital is triage, assigning the patient to an appropriate department based on symptoms. In \method, we employ a router agent that plays the role of a general practitioner. Its parameters are our optimization target $\pi_\theta$, which decides when to invoke multi-agent collaboration.

Concretely, we instantiate a local MLLM as the router. It generates an initial assessment $\tilde{y}$ and appends a certainty token after $\tilde{y}$ for each medical query:
\begin{equation}
\tilde{y} \in \mathcal{Y} \cup \{\texttt{CERTAIN} \mid \texttt{UNCERTAIN}\},
\end{equation}
where the certainty token indicates whether the router is sufficiently certain about its initial assessment $\tilde{y}$. As shown in Fig.~\ref{fig:overview}, if the router is confident (outputs \texttt{CERTAIN}), \method directly adopts $\tilde{y}$ as the final answer and skips the expensive multi-agent collaboration. Otherwise, the router recruits a small subset of expert agents to perform collaborative reasoning. In this case, a moderator agent gathers the analysis reports from the recruited experts and makes the final decision. This discussion‑and‑decision workflow follows prior work~\cite{tang2024medagents, Xia2026MMedAgentRL}, with the designed prompts and role descriptions provided in Appendix~\ref{sec:prompts}. Thus, the workflow of \method is fully adaptive, operating in either single‑agent or multi‑agent mode depending on the given query.

\paragraph{Reward-Shaped Optimization.}
We employ Group Relative Policy Optimization (GRPO)~\cite{shao2024deepseekmath} to jointly optimize answer correctness, uncertainty calibration, reasoning quality, and output format compliance. First, we define a format reward:
\begin{equation}
R_{\mathrm{FMT}}(o_i) =
\mathbb{I}\!\left[o_i \text{ strictly matches the required output schema}\right],
\end{equation}
which enforces structural constraints (e.g., certainty level, reasoning chain, final answer) on the generated response $o_i$.

Next, we introduce two customized rewards: a \emph{Certainty-aware Accuracy Reward (CAR)} and a \emph{Reasoning-quality Reward (RR)}. Let $y_i$ be the answer extracted from $o_i$, $c_i$ the certainty decision ($c_i=1$ for \texttt{CERTAIN}, $0$ for \texttt{UNCERTAIN}), and $y^\star$ the ground‑truth answer. We define CAR as:
\begin{equation}
R_{\mathrm{CAR}}(o_i)
=
\mathbb{I}(y_i = y^\star)\cdot\big[c_i + (1-c_i)\alpha\big]
-
\mathbb{I}(y_i \neq y^\star)\cdot c_i \beta,
\label{eq:car}
\end{equation}
where $\alpha \in (0,1)$ gives reduced credit to correct but uncertain outputs, and $\beta > 0$ penalizes confident mistakes. This reward induces a selective prediction objective with a \emph{reject option}, encouraging the model to output \texttt{UNCERTAIN} when its direct answer is likely wrong. For RR, we use a frozen Vision‑Language Model (VLM) fine‑tuned on our dataset:
\begin{equation}
R_{\mathrm{RR}}(o_i) = f_{\mathrm{VLM}}(x,q,r_i,y_i),
\end{equation}
where $r_i$ is the generated reasoning trace. The combined raw reward is:
\begin{equation}
\bar{R}_i
=
\lambda_{\mathrm{fmt}} R_{\mathrm{FMT}}(o_i)
+
\lambda_{\mathrm{car}} R_{\mathrm{CAR}}(o_i)
+
\lambda_{\mathrm{rr}} R_{\mathrm{RR}}(o_i),
\end{equation}
with coefficients $\lambda_{\mathrm{fmt}}, \lambda_{\mathrm{car}}, \lambda_{\mathrm{rr}}$ controlling the relative importance of format compliance, answer correctness/certainty, and reasoning quality.

Following standard GRPO, we sample a group of $G$ candidate outputs $\{o_i\}_{i=1}^{G}$ for each input $(x,q)$ and standardize the rewards within the group:
\begin{equation}
A_i = \frac{\bar{R}_i - \mu_R}{\sigma_R},
\end{equation}
where $\mu_R$ and $\sigma_R$ are the group mean and variance. The standardized advantage $A_i$ measures how much better a response is relative to its group under our rewards. For simplicity, we omit the KL constraint, and the objective can be expressed in an approximate form as follows:
\begin{equation}
\begin{aligned}
-\mathbb{E}_{(x,q)\sim\mathcal{D},\,\{o_i\}_{i=1}^{G}\sim\pi_{\theta_{\mathrm{old}}}}
\Bigg[
\frac{1}{G}\sum_{i=1}^{G}\frac{1}{|o_i|}
\sum_{t=1}^{|o_i|}
\min\Big(
r_{i,t}(\theta)A_{i,t},
\operatorname{clip}\big(
r_{i,t}(\theta),1-\epsilon,1+\epsilon
\big)A_{i,t}
\Big)
\Bigg]\\,
\end{aligned}
\end{equation}
where $\pi_{\mathrm{ref}}$ is a frozen reference policy, and $\beta_{\mathrm{KL}}$ controls the deviation from the reference. This objective jointly encourages correct answers, calibrated uncertainty, high‑quality reasoning traces, and strict format adherence.

\paragraph{Theoretical Justification.}
We provide a decision-theoretic interpretation of our certainty-aware
design, together with its implications for optimization under GRPO. Let $z \in \{0,1\}$ denote whether the router's direct answer is correct, and let
\begin{equation}
p(x,q) = \Pr(z=1\mid x,q)
\end{equation}
denote the conditional correctness probability of the direct answer.
In the standard setting without deferral, the model always produces an
answer, yielding the expected risk under $0$-$1$ loss:
\begin{equation}
R_{\text{no\_rej}}(x,q) = 1-p(x,q).
\end{equation}

Our formulation introduces an additional decision: the router may either
answer directly or declare \texttt{UNCERTAIN}, which triggers multi-agent
collaboration. This can be interpreted as a form of
\emph{deferred decision-making}, where uncertain cases are routed to a
collaborative reasoning process. To characterize the reward-induced
routing preference, consider an idealized decision setting in which
$p(x,q)$ is calibrated and the router selects the action maximizing its
expected certainty-aware reward. Under the certainty-aware reward in
Eq.~(\ref{eq:car}), the expected utilities of the two actions can be
written as
\begin{equation}
U_{\text{ans}}
=
p(x,q) - (1-p(x,q))\beta,
\qquad
U_{\text{def}}
=
p(x,q)\alpha,
\label{eq:u_ans}
\end{equation}
where $\alpha \in (0,1)$ gives partial credit to correct but deferred
outputs, and $\beta>0$ penalizes overconfident errors. Comparing the two
utilities gives the reward-induced Bayes-optimal threshold:
\begin{equation}
U_{\text{ans}} \geq U_{\text{def}}
\quad\Longleftrightarrow\quad
p(x,q) \geq
\tau_{\mathrm{Bayes}},
\qquad
\tau_{\mathrm{Bayes}}
=
\frac{\beta}{1-\alpha+\beta}.
\label{eq:decision}
\end{equation}
Thus, the certainty-aware reward induces a non-trivial utility boundary: direct answering is preferred
when the estimated correctness probability exceeds $\tau_{\mathrm{Bayes}}$, whereas deferral is preferred otherwise. The detailed derivation of this reward-induced threshold is provided in
Appendix~\ref{sec:derivation}. Importantly, $\tau_{\mathrm{Bayes}}$ characterizes the
\emph{reward-induced theoretical threshold} rather than the exact
decision boundary of the GRPO-trained router. In practice, GRPO learns a
parameterized policy $\pi_{\theta}(a\mid x,q)$, and the optimized router
is given by
\begin{equation}
\theta^\star
\approx
\arg\max_{\theta}
J_{\mathrm{GRPO}}(\theta;\alpha,\beta),
\end{equation}
with the actual routing decision determined by
\begin{equation}
a_{\theta^\star}(x,q)
=
\arg\max_{a\in\{\mathrm{ANS},\mathrm{DEF}\}}
\pi_{\theta^\star}(a\mid x,q).
\label{eq:actual_policy}
\end{equation}
Accordingly, the actual learned routing boundary is implicitly defined
by
\begin{equation}
\pi_{\theta^\star}(\mathrm{ANS}\mid x,q)
=
\pi_{\theta^\star}(\mathrm{DEF}\mid x,q).
\label{eq:policy_boundary}
\end{equation}
The derivation above provides a principled
decision-theoretic interpretation of how $\alpha$ and $\beta$ shape the
relative utility of direct answering and deferral. In practice, the
resulting routing behavior is learned implicitly through GRPO and depends
not only on $\alpha$ and $\beta$, but also on the training data,
optimization dynamics, and the learned policy.

From an optimization perspective, a standard reward
$R_{\text{base}}
=
\mathbb{I}(y=y^\star)
+
\lambda_{\text{fmt}}\mathbb{I}(\text{format})$
cannot distinguish overconfident errors from calibrated deferral, since
both are treated identically with respect to answer correctness, which optimizes in a more narrow landscape and cannot penalize over-confident or over-conservative cases.

\begin{figure}
    \centering
    \includegraphics[width=\linewidth]{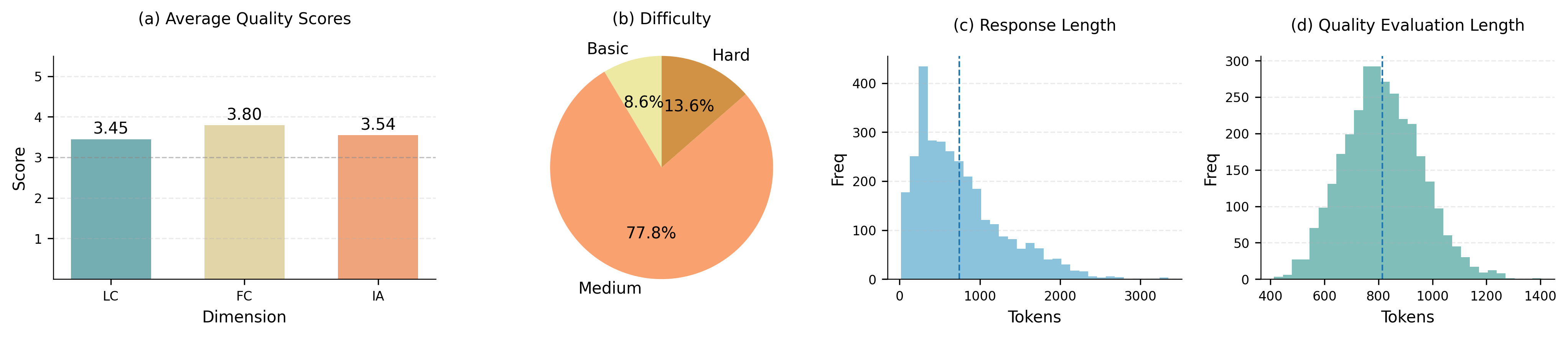}
    \caption{Statistics of our constructed \dataset. (a)-(d) respectively present the average quality score, distribution of difficulty level, response length and quality assessment length.}
    \label{fig:stat}
\end{figure}

\subsection{\dataset: Medical Reasoning-Centric Dataset}
To improve the reasoning capability of local MLLMs, we construct \dataset, a reasoning‑centric multimodal medical dataset. As shown in Fig.~\ref{fig:stat}, besides the medical question and image (e.g., MRI, CT, pathology), each sample in \dataset additionally comprises (i) a gold response with a step‑by‑step reasoning chain, (ii) a difficulty label, and (iii) a quality score along with a justification. Unlike prior MAS that only supervise final answers~\cite{Xia2026MMedAgentRL}, \dataset explicitly provides the reasoning process, offering a more informative supervision signal for multimodal reasoning. Examples are shown in Appendix~\ref{sec:examples}.

\paragraph{Data Collection.}
As illustrated in Fig.~\ref{fig:overview}, we collect question‑image pairs from the training splits of four public benchmarks: Slake~\cite{liu2021slake}, PMC-VQA~\cite{Zhang2024PMCVQA}, VQA-RAD~\cite{lau2018dataset}, and PathVQA~\cite{he2020pathvqa}. For each sample $(x,q)$, we query a set of strong candidate MLLMs (\ie GPT‑5.4~\cite{openai2024gpt4}, Gemini 3 Pro~\cite{geminiteam2024}, Qwen3‑Max~\cite{bai2025qwen3vl}, and Kimi K2.5~\cite{team2025kimi}) and require each response to include a structured, step‑by‑step reasoning chain together with a final answer. Among the candidate responses, we retain only those whose final answers match the ground‑truth label. This first‑stage filtering ensures that the dataset contains reliable reasoning traces instead of fluent but incorrect explanations.

\paragraph{Annotation and Quality Control.}
After correctness filtering, we perform a second‑stage quality control using GPT‑5.4 as an independent judge $J$. Instead of a single scalar score, the judge evaluates each response along three dimensions: \emph{Logic Consistency}, \emph{Factual Correctness}, and \emph{Image Alignment}. Formally, for a candidate response $r$, the judge produces a score vector
\begin{equation}
\mathbf{s} = \big(s_{\mathrm{logic}},\, s_{\mathrm{fact}},\, s_{\mathrm{img}}\big)
= J(x,q,r),
\end{equation}
where the score $s$ is in range of $[1,5]$. We adopt a dimension‑wise filtering strategy and retain a response only if it satisfies all quality thresholds (denoted by $\gamma$):
\begin{equation}
s_{\mathrm{logic}} \ge \gamma_{\mathrm{logic}}, \quad
s_{\mathrm{fact}} \ge \gamma_{\mathrm{fact}}, \quad
s_{\mathrm{img}} \ge \gamma_{\mathrm{img}}.
\end{equation}
This multi‑criteria filtering ensures that retained reasoning traces are not only correct in final answers but also logically coherent, factually accurate, and well‑grounded in visual evidence. Compared to single‑score filtering, this approach avoids cases where strong performance in one dimension compensates for deficiencies in others, yielding more reliable supervision. Subsequently, as shown in Fig.~\ref{fig:stat}, the average quality score of \dataset reaches around 3.60.

\paragraph{Difficulty Annotation.}
Beyond quality control, we further leverage the judge model to annotate the difficulty of each sample. Specifically, the judge assigns a difficulty label $d$ based on both question complexity and visual complexity:
\begin{equation}
d = J(x,q,r) \in \{\texttt{basic}, \texttt{medium}, \texttt{hard}\}.
\end{equation}
The difficulty is determined by factors such as the number of reasoning steps, the level of domain knowledge required, and the ambiguity or complexity of visual evidence. In particular, \texttt{basic} samples typically involve direct visual recognition, while \texttt{medium} and \texttt{hard} samples require multi‑step reasoning, cross‑region inference, or integration of domain‑specific knowledge. This hierarchical annotation enables curriculum learning, allowing the model to progressively learn from simple to complex reasoning patterns. Detailed prompts are provided in Appendix~\ref{sec:prompts}.

After the two‑stage filtering and hierarchical annotation, \dataset contains 3,477 multimodal medical QA samples with structured reasoning chains, which serves as a hierarchical MedVQA dataset specifically designed for reasoning‑centric multimodal medical learning.

\section{Experiments}
\subsection{Setup}
We evaluate our method on four widely-used public MedVQA datasets: Slake~\cite{liu2021slake}, PathVQA~\cite{he2020pathvqa}, VQA-RAD~\cite{lau2018dataset}, PMC-VQA~\cite{Zhang2024PMCVQA}, which cover diverse imaging modalities and clinical scenarios. 
We compare our method with a series of single agent and multi-agent methods as follows: 
\begin{itemize}
    \item \textbf{Single Agent.} We adopt different types and sizes of representative open-source MLLMs to initialize a single agent with basic prompts to serve as a reference, including Qwen3-VL series~\cite{bai2025qwen3vl}, LLaVA~\cite{liu2023visual}, Ministral 3 series~\cite{liu2026ministral}, and the specialized medical models, the BiomedGPT~\cite{zhang2024generalist}, MedFlamingo~\cite{moor2023medflamingo}, HuatuoGPT-Vision~\cite{chen2024huatuogptvisioninjectingmedicalvisual}, and LLaVA-Med~\cite{li2023llavamed}.
    \item \textbf{Multi-agent Collaboration.} We also implement the latest and state-of-the-art MAS methods: MDAgents~\cite{kim2024mdagents}, MedAgent~\cite{tang2024medagents}, MOA~\cite{wang2024mixture}, SMOA~\cite{li2025smoa}, Reconcile~\cite{chen2024reconcile}, and ColaCare~\cite{wang2025colacare}.
\end{itemize}
More information about the datasets, implementation details and evaluation metrics can be found in Appendix~\ref{sec:exp_details}.

\begin{table}[!t]
\vspace{-2mm}
\centering
\scriptsize
\caption{Accuracy comparison with state-of-the-art baseline models on Slake, PathVQA, VQA-RAD and PMC-VQA datasets under multi-choice and free-form questions.}
\vspace{0.5em}
\label{tab:exp}
\resizebox{\textwidth}{!}{%
\renewcommand{\arraystretch}{1.1}
\begin{tabular}{@{}llccccccc@{}}
\toprule
\multirow{2}{*}{\textbf{Category}} & \multirow{2}{*}{\textbf{Methods}} & \multicolumn{4}{c}{\textbf{Multiple-Choice Questions (MC)}} & \multicolumn{3}{c}{\textbf{Free-form Questions (FF)}} \\
\cmidrule(lr){3-6} \cmidrule(rl){7-9}
& & \makecell{Slake} & \makecell{PathVQA} & \makecell{VQA-RAD} & \makecell{PMC-VQA} & \makecell{Slake} & \makecell{PathVQA} & \makecell{VQA-RAD} \\
\midrule

\multirow{4}{*}{\makecell[l]{Single agent \\ (Specialized model)}}
& BiomedGPT   & 53.5\stddev{0.3} &	57.0\stddev{0.7}&	52.0\stddev{0.2} &	41.5\stddev{0.7}  & 54.7\stddev{0.8} & 39.5\stddev{2.2}	& 54.8\stddev{3.4}\\
& Med-Flamingo &61.5\stddev{1.5} & 58.5\stddev{3.0} & 54.5\stddev{1.9} & 38.0\stddev{2.0} & 57.9\stddev{1.1} &	34.7\stddev{2.8}	& 51.5\stddev{4.2} \\
& LLaVA-Med    & 59.0\stddev{0.1}	& 52.5\stddev{0.1} &	60.5\stddev{0.4} & 	36.0\stddev{1.3} & 59.6\stddev{0.4}	& 48.5\stddev{0.5} & 65.8\stddev{1.7}       \\
& HuatuoGPT-Vision  & 69.5\stddev{1.8} & 59.0\stddev{1.4} & 64.0\stddev{1.6} & 41.0\stddev{2.2}  & 59.6\stddev{1.8}	& 45.9\stddev{1.6} & 63.8\stddev{0.6} \\
\hdashline

\multirow{7}{*}{\makecell[l]{Single agent \\ (Local models)}}
& LLaVA-v1.6-7B     & 65.0\stddev{0.5}	& 52.0\stddev{1.2}	& 54.0\stddev{0.7} & 32.0\stddev{2.1} & 52.4\stddev{2.8} & 31.3\stddev{0.7} & 50.8\stddev{2.9}\\
& Minstral-3-3B-Instruct & 68.0\stddev{1.2}	& 51.0\stddev{0.4} & 55.5\stddev{1.1} & 45.0\stddev{2.3} & 67.2\stddev{1.2}	& 33.3\stddev{1.9}	& 45.7\stddev{0.3}\\
& Minstral-3-8B-Instruct & 70.0\stddev{0.8} & 63.0\stddev{1.7} & 57.0\stddev{1.5} & 38.5\stddev{1.1} & 77.3\stddev{1.0}	& 41.7\stddev{0.3} & 62.9\stddev{0.7} \\
& Qwen3-VL-2B-Instruct    & 76.0\stddev{0.5} & 58.5\stddev{1.1} & 66.7\stddev{1.0} & 30.2\stddev{1.3} & 73.6\stddev{2.2} & 40.5\stddev{1.6} & 58.0\stddev{0.1}  \\
& Qwen3-VL-4B-Instruct    & 76.1\stddev{1.4}& 64.6\stddev{2.1}&70.8\stddev{1.8}&35.3\stddev{2.8}&75.4\stddev{0.8}&44.3\stddev{2.5}& 62.6\stddev{4.2} \\
& Qwen3-VL-8B-Instruct    & 71.9\stddev{2.3}&59.0\stddev{0.4}&64.2\stddev{0.5}&50.2\stddev{2.4} & 73.3\stddev{0.7}&45.8\stddev{3.0}&60.3\stddev{5.0} \\
& Qwen3-VL-8B-Thinking     & 73.6\stddev{0.8}&59.0\stddev{1.0}&60.5\stddev{2.3}&33.8\stddev{1.0} & 67.6\stddev{2.4}&49.6\stddev{0.8}&53.6\stddev{1.8} \\

\hdashline
\multirow{6}{*}{\makecell[l]{Multi-agent \\ (Local models)}}
& MedAgents    & 72.0\stddev{1.4}	& 73.7\stddev{2.2}	& 72.0\stddev{0.1} & \underline{52.5}\stddev{0.1} & 74.0\stddev{0.8} & 48.8\stddev{1.6} & \underline{67.0}\stddev{3.0} \\
& MDAgents     & 70.2\stddev{1.2} & \underline{75.6}\stddev{1.0} & 68.6\stddev{2.0} & 50.6\stddev{1.0} & 69.1\stddev{1.9}	& 46.6\stddev{0.6}	& 57.4\stddev{4.3}\\
& MOA     & \underline{79.0}\stddev{1.0} & 69.2\stddev{2.1} & 71.5\stddev{0.5} & 51.5\stddev{1.1} & 72.9\stddev{0.7} & \underline{52.6}\stddev{4.0} & 62.3\stddev{2.7} \\
& Reconcile    & 72.8\stddev{1.0} & 64.2\stddev{1.2} & 66.0\stddev{0.5} & 49.0\stddev{1.0} & 58.1\stddev{1.0} & 43.1\stddev{1.8} & 57.5\stddev{4.3}\\
& SMOA     & 68.0\stddev{1.1} & 62.2\stddev{3.7} & 63.6\stddev{1.2} & 49.3\stddev{2.6} & 60.3\stddev{1.0} & 38.9\stddev{1.1} & 63.0\stddev{1.5}\\
& ColaCare     & 71.3\stddev{1.5} & 60.8\stddev{2.0} & 65.3\stddev{0.6}	& 50.0\stddev{0.8} & 60.5\stddev{1.8} & 45.2\stddev{1.0} & 53.8\stddev{2.6}\\

\hdashline

\multirow{2}{*}{\makecell[l]{Adaptive agent \\ (Local models)}}
& \method (Ours) & 76.7\stddev{1.0}	& 68.6\stddev{1.3}	& \underline{72.5}\stddev{0.5} & 48.5\stddev{2.1} & \underline{78.2}\stddev{1.3} & 48.7\stddev{0.2} & 64.1\stddev{3.2} \\
& \method* (Ours) & \textbf{87.0}\stddev{0.5} & \textbf{77.0}\stddev{1.1}	& \textbf{74.0}\stddev{1.6}	& \textbf{72.5}\stddev{1.0} & \textbf{84.3}\stddev{2.5} & \textbf{58.5}\stddev{3.5} & \textbf{69.4}\stddev{3.7} \\

\bottomrule
\end{tabular}
}
\captionsetup{justification=raggedright,singlelinecheck=false,font=tiny}

\caption*{
\textit{\textbf{Note:}} MC: Multiple-choice question answering; FF: Free-form (open-ended) question answering. \textbf{Bold} indicates the best performance, and \underline{Underlined} indicates the second-best performance per column (dataset and task). All scores are reported as mean\stddev{standard deviation} by testing 5 times independently. For fair comparison, we use Qwen3-VL-4B-Instruct as the base model for MAS approaches. \method* denotes our method after fine-tuning on \dataset.
}
\vspace{-2mm}
\end{table}

\subsection{Results Analysis}
\paragraph{Quantitative Comparison.}
As shown in Table~\ref{tab:exp},  without any fine-tuning, \method already achieves competitive performance among local multi-agent approaches. For example, on FF tasks, \method attains 78.2\% on Slake, outperforming the best local baseline (77.3\% from Minstral-3-8B-Instruct) and surpassing most multi-agent methods. On PathVQA and VQA-RAD, it achieves 48.7\% and 64.1\%, respectively, remaining competitive with dense multi-agent frameworks such as MOA (52.6\%, 62.3\%) and MedAgents (48.8\%, 67.0\%). On MC tasks, \method also performs strongly, reaching 76.7\% on Slake and achieving the second-best result on VQA-RAD (72.5\%), while maintaining reasonable performance on the more challenging PMC-VQA (48.5\%). 

After fine-tuning on \dataset, \method* further achieves substantial improvements across all benchmarks. On MC tasks, \method* reaches 87.0\% on Slake, 77.0\% on PathVQA, 74.0\% on VQA-RAD, and 72.5\% on PMC-VQA, exceeding the best baselines by +8.0\%, +1.4\%, +2.0\%, and +20.0\% points, respectively. On FF tasks, it achieves 84.3\% on Slake, 58.5\% on PathVQA, and 69.4\% on VQA-RAD, establishing new state-of-the-art results among local models. Notably, the largest gains are observed on PMC-VQA and PathVQA, which involve more complex reasoning, highlighting the effectiveness of our reasoning-centric training and reward design. Overall, these results show that while \method provides a strong and efficient zero-shot solution, \method* further unlocks its potential through reasoning supervision and uncertainty-aware optimization, leading to consistently superior performance. 

\begin{table}[!t]
\centering
\scriptsize
\caption{Accuracy comparison with representative multi-agent collaboration methods after fine-tuning on our curated \dataset dataset using identical fine-tuning protocol.}
\label{tab:matched}
\scriptsize
\begin{tabular}{@{}lccccccc@{}}
\toprule
\multirow{2}{*}{\textbf{Method}} & \multicolumn{4}{c}{\textbf{MC}} & \multicolumn{3}{c}{\textbf{FF}} \\
\cmidrule(lr){2-5} \cmidrule(rl){6-8}
& Slake & PathVQA & VQA-RAD & PMC-VQA & Slake & PathVQA & VQA-RAD \\
\midrule
MOA* & 85.6\stddev{1.8} & 70.0\stddev{1.4} & \textbf{75.5}\stddev{1.1} & 65.3\stddev{0.6} & 79.0\stddev{0.9} & \underline{56.0}\stddev{0.4} & 61.3\stddev{0.7} \\

MDAgent* & \underline{86.7}\stddev{0.9} & \underline{75.3}\stddev{1.7} & 71.9\stddev{2.1} & \underline{67.7}\stddev{1.9} & \underline{80.9}\stddev{0.8} & 50.5\stddev{2.2} & 53.9\stddev{5.0} \\

\method* (Ours) & \textbf{87.0}\stddev{1.0} & \textbf{77.0}\stddev{1.1} & \underline{74.0}\stddev{1.6} & \textbf{72.5}\stddev{1.0} & \textbf{84.3}\stddev{2.5} & \textbf{58.5}\stddev{3.5} & \textbf{69.4}\stddev{3.7} \\
\bottomrule
\end{tabular}

\end{table}

To disentangle the accuracy gains of \method* from the effect of fine-tuning on \dataset itself, we additionally fine-tune two representative baselines, MOA and MDAgent, with the identical \dataset corpus and training pipeline (replacing the certainty-aware reward with the standard accuracy reward, since neither models a certainty token). Table~\ref{tab:matched} shows that \dataset consistently benefits all methods, yet \method* still achieves the best overall performance after controlling for identical post-training, indicating that uncertainty-aware routing provides complementary benefits beyond domain-specific fine-tuning alone.

\begin{table}[]
    \centering
    \scriptsize
    \caption{Reasoning quality comparison. LC, FC, and IA respectively represent the Logic Consistency, Factual Correctness and Image Alignment.}
    \vspace{0.25em}
    \resizebox{0.9\linewidth}{!}{
    \begin{tabular}{cccccc|cc}
    \hline
    \textbf{Score} & \textbf{MDAgent} & \textbf{MedAgent} & \textbf{SMOA} & \textbf{MOA} & \textbf{ColaCare} & \method & \method* \\
    \hline
    LC     &  2.89 & 3.31 & \underline{3.48} & 3.42 & 3.28 & 3.24 & \textbf{3.55} \\
    FC    &  2.80 & 3.11 & \underline{3.37} & 3.12 & 3.20 & 2.96 & \textbf{3.38} \\
    IA & 2.56 & 3.19 & 3.29 & \underline{3.40} & 3.18 & 3.14 & \textbf{3.57}\\
    \hdashline
    Avg. & 2.75 & 3.20 & \underline{3.38} & 3.31 &  3.22 & 3.11 & \textbf{3.50} \\
    \hline
    \end{tabular}}
    \label{tab:quality}
\end{table}

\paragraph{Qualitative Comparison.}
We randomly select 100 samples, covering both FF and MC questions for reasoning quality comparison. Similarly, we employ the GPT-5.4 as the judge model, and we evaluate the generated reasoning along three dimensions: Logic Consistency (LC), Factual Correctness (FC), and Image Alignment (IA). The prompts for the judge model can be found in Appendix~\ref{sec:prompts}. The results are reported in Table~\ref{tab:quality}. It can be seen that \method* achieves the best performance across all dimensions, with an average score of 3.50, outperforming prior multi-agent methods such as SMOA (3.38) and MOA (3.31). In particular, \method* shows strong improvements in Image Alignment (3.57 vs. 3.40 of MOA) and Logic Consistency (3.55 vs. 3.48 of SMOA), indicating that our method generates reasoning that is better grounded in visual evidence while maintaining coherent logical structure. Comparing \method and \method*, we observe consistent gains across all metrics (Avg.: 3.11 $\rightarrow$ 3.50), validating the effectiveness of our GRPO training with certainty-aware and reasoning rewards. More validations and analysis can be found in Appendix~\ref{sec:add}.

\begin{figure}
    \centering
    \includegraphics[width=\linewidth]{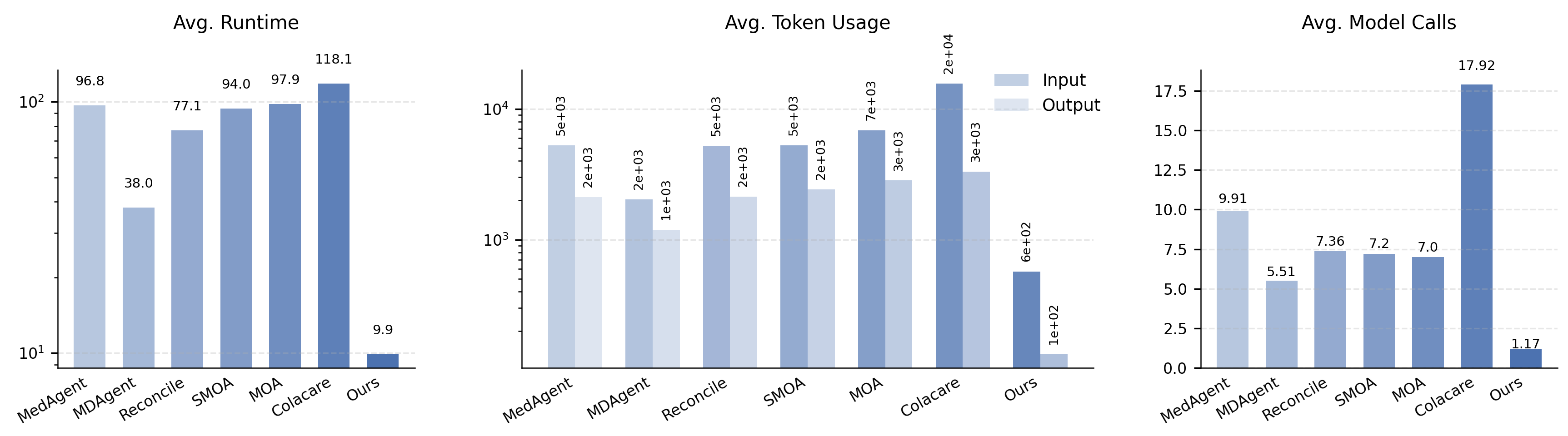}
    \caption{Efficiency comparison of \method and other MAS methods. We compare the average runtime (second), average I/O token usage and the average model calls (times) per question.}
    \label{fig:efficiency}
\end{figure}

\paragraph{Efficiency Analysis.}
We evaluate efficiency of existing MAS approaches in terms of latency, token usage, and communication overhead on 100 randomly-sampled questions. From Fig.~\ref{fig:efficiency}, it has been shown that \method is significantly more efficient than other approaches, requiring less than 10 seconds for every query on average, corresponding to a 3.8$\times$ speedup over the fastest baseline (\ie MDAgents) and up to 11.9$\times$ over dense collaboration methods (\ie ColaCare). In addition, \method merely consumes around 600 input tokens and 100 output tokens per query, reducing communication cost by over 72\% and 88\%, respectively. It also requires only 1.17 model calls per question, far less than 5.51--17.92 in prior methods. These results verify that selective, uncertainty-aware routing substantially reduces computation and communication overhead.

\begin{table}[]
\centering
\scriptsize
\caption{Ablation study results on MC and FF questions.}
\label{tab:ablation}
\resizebox{0.85\linewidth}{!}{
\begin{tabular}{lccccccc}
\toprule
\multirow{2}{*}{Model} & \multicolumn{4}{c}{MC} & \multicolumn{3}{c}{FF} \\
& Slake & PathVQA & VQA-RAD & PMC-VQA & Slake & PathVQA & VQA-RAD \\
\midrule
\multicolumn{8}{c}{\textbf{Single agent}} \\
 & 76.1\stddev{1.4} & 64.6\stddev{2.1} & 70.8\stddev{1.8} & 35.3\stddev{2.8} & 75.4\stddev{0.8} & 44.3\stddev{2.5} & 62.6\stddev{4.2} \\  
\hdashline
\multicolumn{8}{c}{\textbf{Multi-agent}} \\
+ Dense collaboration & 78.5\stddev{0.3} & 66.7\stddev{2.4} & 71.3\stddev{2.2} & 50.1\stddev{0.5}& 75.7\stddev{2.8} & 51.2\stddev{1.7} & 64.5\stddev{1.7}\\
+ Uncertainty routing & 76.7\stddev{1.0} & 68.6\stddev{1.3} & 72.5\stddev{0.5} & 48.5\stddev{2.1} & 78.2\stddev{1.3} & 48.7\stddev{0.2} & 64.1\stddev{3.2}\\
\quad + Certain. reward &  \underline{85.5}\stddev{1.0} &  \underline{75.0}\stddev{0.9} & \underline{73.1}\stddev{1.2} & \underline{69.5}\stddev{1.1} & \underline{82.2}\stddev{1.8} & \underline{56.8}\stddev{1.2} & \underline{68.7}\stddev{2.6}\\
\qquad + Reason. reward & \textbf{87.0}\stddev{0.5} & \textbf{77.0}\stddev{1.1} & \textbf{74.0}\stddev{1.6} & \textbf{72.5}\stddev{1.0} & \textbf{84.3}\stddev{2.5} & \textbf{58.5}\stddev{3.5} & \textbf{69.4}\stddev{3.7} \\
\bottomrule
\end{tabular}
}
\end{table}

\subsection{Ablation Study}

\paragraph{Effect of Agent Collaboration.} From Table~\ref{tab:ablation}, the dense collaboration improves MC accuracy on PMC-VQA from 35.3\% to 50.1\% (+14.8\%) and PathVQA from 64.6\% to 66.7\%, while also bringing gains on FF tasks such as PathVQA (44.3\% to 51.2\%). Replacing dense collaboration with uncertainty-driven routing achieves comparable or even better performance without any additional training. Notably, uncertainty routing outperforms dense collaboration on PathVQA (66.7\% to 68.6\%) and VQA-RAD (71.3\% to 72.5\%) in MC, and achieves higher FF accuracy on Slake (75.7\% to 78.2\%). These results confirm that multi-agent collaboration should be \emph{selective rather than universal}: many queries can be reliably solved by a single agent, while only a subset of complex cases benefit from the agent collaboration.

\paragraph{Effect of Agent Learning.} Training with GRPO using our proposed Certainty-Aware Accuracy Reward (CAR) further boosts performance across all benchmark. For instance, MC accuracy on PMC-VQA improves significantly from 48.5\% to 69.5\%, and PathVQA from 68.6\% to 75.0\%, while FF performance on Slake increases from 78.2\% to 82.2\%. Furthermore, incorporating the Reasoning-quality Reward (RR) leads to additional gains, particularly on FF tasks where reasoning quality is critical (\eg PathVQA: 56.8\% to 58.5\%, Slake: 82.2\% to 84.3\%), and also improves MC performance (\eg PMC-VQA: 69.5\% to 72.5\%). These results show that the combination of CAR and RR enables the model to both make calibrated routing decisions and generate higher-quality reasoning, resulting in the best overall performance.

\section{Conclusion}

In this paper, we present \method, an adaptive and efficient multi-agent framework for medical multimodal reasoning that leverages uncertainty-driven sparse routing to enable selective collaboration. To support effective training, we construct \dataset, a reasoning-centric MedVQA benchmark with hierarchical difficulty, and propose a unified reward design that jointly models correctness, uncertainty, reasoning quality, and structural consistency. Experiments on multiple benchmarks show that \method achieves state-of-the-art performance among local models while significantly reducing computational overhead and preserving data privacy. These results highlight the potential of fully local agent systems with reasoning-centric fine-tuning for building practical, reliable, and deployable medical AI systems.

\section*{Acknowledgment}
This paper was supported in part by the National Key R\&D Program of China under Grant 2022YFA1008300, in part by the National Natural Science Foundation of China under Grant 12471308, and in part by the Excellent Young Scholars of Shenzhen under Grant RCYX20231211090247060.

\bibliography{reference}
\bibliographystyle{plain}

\newpage


\appendix

\begin{center}
    \huge \textbf{Appendix}
\end{center}
\vspace{1cm}

  \begin{itemize}
      \item \S\ref{sec:exp_details} Implementation Details
          \begin{itemize}
              \item Training Details
              \item Hyper-Parameters
              \item Datasets
              \item Evaluation Metrics
          \end{itemize}
      \item \S\ref{sec:derivation} Derivation of the Reward-Induced Decision Threshold
      \item \S\ref{sec:add} Additional Analysis
          \begin{itemize}
              \item Routing Validation
              \item Risk--Coverage Analysis
              \item False-Certainty Safety Analysis
              \item Statistical Significance
              \item Extra Judge Evaluation
          \end{itemize}
      \item \S\ref{sec:traces_correction} Examples of \method
          \begin{itemize}
              \item Example 1: Pathology Question on Inflammatory Infiltrate Characterization
              \item Example 2: Multiple-Choice Question on Muographic Map Interpretation
              \item Discussion
          \end{itemize}
      \item \S\ref{sec:examples} Showcases of \dataset
      \item \S\ref{sec:prompts} Prompts Design
          \begin{itemize}
              \item Prompts for Building \dataset
              \item Prompts for Building \method
          \end{itemize}
  \end{itemize}
  \vspace{0.3cm}

\newpage
\section{Implementation Details}
\label{sec:exp_details}
\paragraph{Training Details.}
We adopt the Qwen3-VL-4B-Instruct~\cite{bai2025qwen3vl} as the foundation model due to the trade-off between model scale and performance, as well as its superior instruction-following capability in multimodal settings. We further post-train the model on our constructed \dataset to enhance its reasoning ability and uncertainty estimation. Specifically, we employ Low-Rank Adaptation (LoRA)~\cite{hu2022lora} to fine-tune all linear layers of the model for both supervised fine-tuning (SFT) and reinforcement learning. During SFT, we train the model for 5 epochs with a learning rate of $1\times 10^{-4}$. For the subsequent GRPO stage, we adopt a smaller learning rate of $1\times 10^{-6}$ and fine-tune for 1 epoch to stabilize policy optimization. We follow a curriculum learning strategy during training. The model is first trained on samples with lower difficulty levels and progressively exposed to more challenging samples, enabling it to gradually acquire complex reasoning capabilities. In the GRPO stage, we focus exclusively on high-difficulty samples to further push the upper bound of the model’s reasoning performance and decision-making ability under uncertainty. For the VLM reward model, we employ the Qwen3-VL-2B-Instruct and fine-tune it on the response-quality pairs offered by \dataset for 10 epochs. During GRPO fine-tuning, it will be frozen and used to output the quality rewards. All experiments are conducted on 8 NVIDIA Tesla H100 GPUs. We implement our training pipeline based on the ms-swift framework. We set the number of generations for GRPO is 6 and the number of gradient accumulation steps is 3.

\paragraph{Hyper-Parameters.} In our experiments, we set the maximum number of specialists to 3 and discussion rounds to 2 for both baselines and our \method. All agents are initialized from Qwen3-VL-4B-Instruct, except for \method*, which uses a version fine-tuned on our \dataset. Unless otherwise noted, our implementation follows the official repository.

\begin{table}[htb]
    \centering
    \caption{Dataset statistics. We compare the modality, size, question type of 4 datasets as well as our constructed \dataset that are used in our work.}
    \resizebox{\linewidth}{!}{
    \begin{tabular}{cccccc}
    \toprule
         &  Slake & PathVQA & VQA-RAD & PMC-VQA & \dataset \\
    \midrule
      Size & 4,717 & 26,516 & 1,778 & 4,000 & 3,477\\
      Answer & \cmark & \cmark & \cmark & \cmark & \cmark \\
      Reasoning & \xmark & \xmark & \xmark & \xmark & \cmark \\
      Difficulty & \xmark & \xmark & \xmark & \xmark & \cmark \\
      Modality   &  Radiology images & Pathology slides & Radiology images & Radiology images & Pathology + Radiology\\
      Question Type & YES/NO/OPEN & YES/NO/OPEN & YES/NO/OPEN & A/B/C/D & YES/NO/A/B/C/D\\
    \bottomrule
    \end{tabular}}
    \label{tab:dataset}
\end{table}


\paragraph{Datasets.} In this work, we employ 4 popular MedVQA datasets as follows:
\begin{itemize}
    \item \textbf{Slake} is a multilingual medical VQA dataset constructed from radiology images, including X-ray, CT, and MRI, with annotations provided in both English and Chinese, and emphasizes visual grounding with well-aligned question–answer pairs \cite{liu2021slake}. 
    \item \textbf{PathVQA} focuses on pathology images, containing high-resolution histopathological slides with questions that require fine-grained visual reasoning and domain-specific knowledge \cite{he2020pathvqa}.
    \item  \textbf{VQA-RAD} is a radiology VQA benchmark built upon clinical radiology images, where questions often involve abnormality detection, modality recognition, and anatomical reasoning; it is relatively small but widely used for evaluating medical reasoning capability \cite{lau2018dataset}. 
    \item   \textbf{PMC-VQA} is a large-scale dataset derived from biomedical literature, consisting of hundreds of thousands of image–question–answer pairs spanning multiple modalities (e.g., radiology, microscopy, charts) \cite{Zhang2024PMCVQA}. 
\end{itemize}
We employ a part of their training set to construct our \dataset and their test set for evaluation, and the detailed statistics of each dataset are summarized in Table~\ref{tab:dataset}. In particular, after quality control, we adopted 746, 721, 707, 1303 samples from Slake, PathVQA, VQA-RAD, and PMC-VQA respectively. 

\paragraph{Evaluation Metrics.}
We adopt two complementary evaluation metrics to assess model performance under different question formats. (i) Accuracy: For multiple-choice questions, we use exact-match accuracy, which measures whether the predicted answer $\hat{y}$ matches the ground-truth label $y^\star$:
\begin{equation}
\text{Accuracy} = \frac{1}{N} \sum_{i=1}^{N} \mathbb{I}(\hat{y}_i = y_i^\star),
\end{equation}
where $\mathbb{I}(\cdot)$ is the indicator function. This metric is suitable for structured questions with a predefined set of candidate answers. (ii) Correctness:
For free-form questions, exact matching is often overly restrictive due to linguistic variability. Therefore, we adopt a semantic correctness metric based on an independent and proprietary MLLM evaluator. Specifically, given a generated answer $\hat{y}$ and ground-truth answer $y^\star$. In practice, we employ the GPT 5.4 owing to its strong text information processing ability. Formally, the evaluator model $\mathcal{E}$ produces a score:
\begin{equation}
\text{Correctness} = \mathcal{E}(\hat{y}, y^\star) \in [0,1],
\end{equation}
which reflects the semantic alignment between the prediction and the reference. 

\section{Derivation of the Reward-Induced Decision Threshold}
\label{sec:derivation}

We provide a derivation of the reward-induced decision threshold
discussed in Sec.~\ref{sec:method}. At the router level, the model
chooses between answering directly and declaring \texttt{UNCERTAIN},
which triggers multi-agent collaboration. We consider an idealized
decision setting in which $p(x,q)$ is calibrated and the router selects
the action with the highest expected certainty-aware reward.

Under the certainty-aware reward in Eq.~(\ref{eq:car}), the expected
utilities of the two actions are
\begin{equation}
U_{\text{ans}}
=
p(x,q) - (1-p(x,q))\beta,
\qquad
U_{\text{def}}
=
p(x,q)\alpha.
\end{equation}

The Bayes-optimal action under this reward selects the action with higher
expected utility:
\begin{equation}
U_{\text{ans}} \geq U_{\text{def}}.
\end{equation}

Substituting the above utilities, we obtain
\begin{equation}
p(x,q) - (1-p(x,q))\beta
\geq
p(x,q)\alpha.
\end{equation}

Rearranging terms gives
\begin{equation}
p(x,q)(1+\beta-\alpha)
\geq
\beta,
\end{equation}
which yields the reward-induced threshold
\begin{equation}
p(x,q)
\geq
\frac{\beta}{1-\alpha+\beta}.
\end{equation}

Therefore, the corresponding Bayes-optimal decision rule under the
idealized calibrated setting is
\begin{equation}
a_{\mathrm{Bayes}}^*(x,q)
=
\begin{cases}
\mathrm{ANS},
&
p(x,q)\geq\tau_{\mathrm{Bayes}},
\\
\mathrm{DEF},
&
p(x,q)<\tau_{\mathrm{Bayes}},
\end{cases}
\qquad
\tau_{\mathrm{Bayes}}
=
\frac{\beta}{1-\alpha+\beta}.
\end{equation}

This threshold should be interpreted as a theoretical utility
indifference point induced by the reward parameters $\alpha$ and
$\beta$, rather than as the exact decision boundary of the
GRPO-trained router. In the actual model, GRPO learns a policy
$\pi_{\theta^\star}(a\mid x,q)$, whose routing boundary is implicitly
determined by
\begin{equation}
\pi_{\theta^\star}(\mathrm{ANS}\mid x,q)
=
\pi_{\theta^\star}(\mathrm{DEF}\mid x,q).
\end{equation}
Consequently, the empirical routing boundary may differ from
$\tau_{\mathrm{Bayes}}$ due to imperfect calibration, finite training
data, optimization dynamics, and the parameterization of the learned
policy.

\section{Additional Analysis}
\label{sec:add}
\paragraph{Routing Validation.}
To verify that the learned router agent can exploit meaningful sample-level uncertainty rather than merely changing the average collaboration rate, we compare \method* against two extreme policies (always-single, always-collaborate) using the same fine-tuned backbone and a random-routing control that replaces the learned router with an independent Bernoulli decision at a matched collaboration rate. Table~\ref{tab:routing}(a) shows \method* achieves the best accuracy on both MC and FF while issuing only 1.2 calls on average, outperforming both extremes; Table~\ref{tab:routing}(b) shows random routing achieves far lower accuracy than learned routing despite a comparable single-agent rate, indicating the router identifies \emph{which} queries benefit from collaboration rather than just how often to invoke it.

\begin{table}[htb]
\centering
\scriptsize
\caption{Routing validation with a matched fine-tuned backbone. (a) Learned routing vs.\ two extreme policies; (b) learned routing vs.\ random routing at a matched collaboration rate.}
\label{tab:routing}
\begin{tabular}{@{}lccc@{}}
\toprule
\textbf{Model} & \textbf{MC} & \textbf{FF} & \textbf{Avg. Calls} \\
\midrule
\multicolumn{4}{c}{\textit{(a) Routing strategy comparison}} \\
\midrule
Single agent* & 76.2 & 65.4 & 1.0 \\
Dense collaboration* & 77.5 & 66.8 & 4.0 \\
\method* (Ours) & \textbf{78.1} & \textbf{68.5} & \textbf{1.2} \\
\midrule
\multicolumn{4}{c}{\textit{(b) Random vs.\ learned routing (\% single-agent)}} \\
\midrule
Random routing ($p=0.5$) & 65.0 (49.5\%) & 62.2 (48.5\%) & 2.0 \\
Uncertainty routing (Ours) & \textbf{78.1} (85.5\%) & \textbf{68.5} (78.5\%) & \textbf{1.2} \\
\bottomrule
\end{tabular}
\end{table}

\begin{wraptable}{r}{0.4\textwidth}
\caption{Risk--coverage analysis before/after certainty-aware GRPO.}
\label{tab:aurc}
\scriptsize
\begin{tabular}{@{}lcc@{}}
\toprule
\textbf{Model} & \textbf{AURC} $\downarrow$ & \textbf{Acc.@100\%} $\uparrow$ \\
\midrule
\method & 0.2669 & 66.0\% \\
\method* & \textbf{0.2334} & \textbf{71.5\%} \\
\bottomrule
\end{tabular}
\end{wraptable}

\paragraph{Risk--Coverage Analysis.}
We further quantify calibration quality via the Area Under the Risk--Coverage Curve (AURC; lower is better). Table~\ref{tab:aurc} shows certainty-aware GRPO reduces overall AURC from 0.2669 to 0.2334 while improving full-coverage accuracy from 66.0\% to 71.5\%, confirming that our designed GRPO fine-tuning improves calibration and accuracy jointly.

\paragraph{False-Certainty Safety Analysis.}
The false-certainty is a typical failure mode in MedVQA. Table~\ref{tab:falsecertain} cross-tabulates the router's certainty token against correctness before and after GRPO: the false-certainty rate (Incorrect+CERTAIN) drops from 28.8\% to 8.6\%, while incorrect predictions correctly flagged \texttt{UNCERTAIN} rise from 5.3\% to 15.7\%, showing that certainty-aware GRPO specifically suppresses overconfident errors and defers the remaining difficult cases to collaboration, which can then correct them (concrete examples can be found in the Appendix~\ref{sec:traces_correction}).

\begin{table}[htb]
\centering
\scriptsize
\caption{Certainty--correctness confusion matrices before/after certainty-aware GRPO.}
\label{tab:falsecertain}
\renewcommand{\arraystretch}{1.25}

\begin{tabular}{l|cc|cc}
\hline
\multirow{2}{*}{\textbf{Correctness}}& \multicolumn{2}{c|}{\textbf{Before GRPO}} & \multicolumn{2}{c}{\textbf{After GRPO}} \\
\cline{2-5}
 & \textbf{CERTAIN} & \textbf{UNCERTAIN} & \textbf{CERTAIN} & \textbf{UNCERTAIN} \\
\hline
Correct & 63.4\% & 2.5\% & 71.5\% & 4.1\% \\
Incorrect & 28.8\% & 5.3\% & 8.6\% & 15.7\% \\
\hline
\end{tabular}
\end{table}



\begin{table}[htb]
\centering

\caption{Additional robustness analyses. (a) Paired significance tests of \method* vs.\ baselines (five runs). (b) Independent judge (Gemini~3~Pro) vs.\ GPT-5.4 for reasoning-quality scores.}
\label{tab:robustness_appendix}
\begin{minipage}{0.48\textwidth}
\centering
\textit{(a) Significance vs.\ baselines}\\[2pt]
\begin{tabular}{@{}lc@{}}
\toprule
\textbf{Comparison} & \textbf{$p$-value} \\
\midrule
vs.\ Reconcile & 0.0025 \\
vs.\ ColaCare & 0.0043 \\
vs.\ MedAgent & 0.0137 \\
vs.\ MDAgent & 0.0341 \\
vs.\ MOA & 0.0860 \\
\bottomrule
\end{tabular}
\end{minipage}
\hfill
\begin{minipage}{0.48\textwidth}
\centering
\textit{(b) Judge comparison}\\
\begin{tabular}{@{}lcc@{}}
\toprule
\textbf{Method} & \textbf{GPT-5.4} & \textbf{Gemini 3 Pro} \\
\midrule
MDAgent & 2.75 & 3.12 \\
MedAgent & 3.20 & 2.94 \\
SMOA & 3.38 & 3.35 \\
MOA & 3.31 & 3.42 \\
ColaCare & 3.22 & 3.57 \\
\method & 3.11 & 3.20 \\
\method* & \textbf{3.50} & \textbf{3.73} \\
\bottomrule
\end{tabular}
\end{minipage}
\end{table}

\paragraph{Statistical Significance.}
We perform paired significance tests using per-question performance of \method* against each baseline across the five runs, reported in Table~\ref{tab:robustness_appendix}(a). The improvements over Reconcile ($p=0.0025$) and ColaCare ($p=0.0043$) are significant at the 1\% level, and the improvements over MedAgent ($p=0.0137$) and MDAgent ($p=0.0341$) are significant at the 5\% level. The comparison with MOA does not reach significance ($p=0.0860$), which is consistent with the relatively small performance gap between the two methods observed in Table~\ref{tab:exp}. Overall, the pairwise comparisons are statistically significant at different salient levels, supporting that the accuracy gains of \method* are statistically salient against most of baseline MAS models.

\paragraph{Extra Judge Evaluation.}
To verify the reasoning quality improvements of our reasoning quality reward across different judge models, we additionally score the same 100 sampled responses with Gemini 3 Pro, a model that is completely independent of both \dataset construction and reward-model training, and report the overall average score (across the Logic Consistency, Factual Correctness, and Image Alignment dimensions used in Table~\ref{tab:quality}) under both judges in Table~\ref{tab:robustness_appendix}(b). The ranking is largely preserved across the two judges, and \method* remains the top-scoring method under both GPT-5.4 (3.50) and Gemini 3 Pro (3.73). Notably, Gemini 3 Pro assigns \method* an even higher score than GPT-5.4 does, and the relative ordering of the multi-agent baselines (\eg ColaCare and MOA scoring higher under Gemini 3 Pro than under GPT-5.4) shifts somewhat but does not change the overall conclusion. This cross-judge consistency indicates that the reasoning-quality advantage of \method* reflects a genuine property of the generated reasoning traces rather than a preference specific to GPT-5.4.

\section{Examples of our \method}
\label{sec:traces_correction}

While the quantitative results in Sec.~\ref{sec:method} and the efficiency analysis in Fig.~\ref{fig:efficiency} demonstrate the efficiency gains of \method on straightforward cases, it is equally important to examine cases where the router's initial assessment is uncertain or incorrect, and where multi-agent collaboration successfully corrects the prediction. In this section, we present detailed inference traces contrasting \method against baseline multi-agent systems (MDAgent, MOA, and MedAgent where available) on two representative difficult cases from our evaluation. These examples illustrate not only that uncertainty-aware routing avoids unnecessary collaboration on easy cases, but also that collaboration is selectively invoked and effective precisely when the initial prediction is ambiguous or erroneous.

\paragraph{Example 1: Pathology Question on Inflammatory Infiltrate Characterization.}

\begin{custommdframed}
\begin{description}[leftmargin=0pt,labelindent=0pt,labelsep=0.5em,itemsep=0.3em]
    \item[\textbf{Question:}] Is the immediate reaction characterized by an inflammatory infiltrate rich in eosinophils, neutrophils, and T cells?
    \item[\textbf{Ground Truth:}] NO
    \item[\textbf{SMART-Med:}] 
        \begin{itemize}[leftmargin=1em,topsep=0pt,itemsep=0pt]
            \item Initial assessment $\to$ certainty=UNCERTAIN $\to$ answer=YES (incorrect)
            \item Router recruits expert: Pathology
            \item Specialist analysis $\to$ Moderator aggregation
            \item Final decision=NO \textbf{(correct)}
            \item Calls=4; tokens\_in/out=1468/355; time=34.75s
        \end{itemize}
    \item[\textbf{MDAgent:}]
        \begin{itemize}[leftmargin=1em,topsep=0pt,itemsep=0pt]
            \item Difficulty=intermediate $\to$ recruit=[Pulmonologist, Allergist/Immunologist, Pathologist]
            \item Specialist analyses and repeated inter-agent consultation
            \item Final decision=YES \textbf{(incorrect)}
            \item Calls=34; tokens\_in/out=8184/4464; time=303.91s
        \end{itemize}
    \item[\textbf{MOA:}]
        \begin{itemize}[leftmargin=1em,topsep=0pt,itemsep=0pt]
            \item Three homogeneous agent analyses $\to$ moderator aggregation
            \item Final decision=NO \textbf{(correct)}
            \item Calls=4; tokens\_in/out=2093/1728; time=123.73s
        \end{itemize}
    \item[\textbf{MedAgent:}]
        \begin{itemize}[leftmargin=1em,topsep=0pt,itemsep=0pt]
            \item Domain classification $\to$ domains=[Rheumatology, Immunology, Allergy]
            \item Domain analyses $\to$ synthesis/consensus
            \item Final decision=NO \textbf{(correct)}
            \item Calls=12; tokens\_in/out=6067/2738; time=183.68s
        \end{itemize}
\end{description}
\end{custommdframed}

\paragraph{Example 2: Multiple-Choice Question on Muographic Map Interpretation.}

\begin{custommdframed}
\begin{description}[leftmargin=0pt,labelindent=0pt,labelsep=0.5em,itemsep=0.3em]
    \item[\textbf{Question:}] What do the empty white stars indicate in the map of the measurement site? \\ 
    \hspace*{1em} A: Selected cross-sections. \\
    \hspace*{1em} B: Locations of lahar observation stations. \\
    \hspace*{1em} C: Selected angular regions. \\
    \hspace*{1em} D: Locations of muographic observation stations.
    \item[\textbf{Ground Truth:}] B
    \item[\textbf{SMART-Med:}]
        \begin{itemize}[leftmargin=1em,topsep=0pt,itemsep=0pt]
            \item Initial assessment $\to$ certainty=UNCERTAIN $\to$ answer=B (correct)
            \item Router recruits expert: General Practice
            \item Specialist analysis $\to$ Moderator aggregation
            \item Final decision=B \textbf{(correct)}
            \item Calls=4; tokens\_in/out=2231/394; time=38.34s
        \end{itemize}
    \item[\textbf{MDAgent:}]
        \begin{itemize}[leftmargin=1em,topsep=0pt,itemsep=0pt]
            \item Difficulty=intermediate $\to$ recruit=[Urologist, Pathologist, Oncologist]
            \item Specialist analyses and inter-agent consultation
            \item Final decision=B \textbf{(correct)}
            \item Calls=13; tokens\_in/out=5705/1851; time=113.65s
        \end{itemize}
    \item[\textbf{MOA:}]
        \begin{itemize}[leftmargin=1em,topsep=0pt,itemsep=0pt]
            \item Three homogeneous agent analyses $\to$ moderator aggregation
            \item Final decision=D \textbf{(incorrect)}
            \item Calls=4; tokens\_in/out=2657/1446; time=81.97s
        \end{itemize}
    \item[\textbf{MedAgent:}]
        \begin{itemize}[leftmargin=1em,topsep=0pt,itemsep=0pt]
            \item Domain classification $\to$ domains=[Field0, Field1, Field2]
            \item Domain analyses $\to$ synthesis/consensus
            \item Final decision=B \textbf{(correct)}
            \item Calls=13; tokens\_in/out=8930/2430; time=164.17s
        \end{itemize}
\end{description}
\end{custommdframed}

\paragraph{Discussion.}
These two examples demonstrate that \method's router recognizes its own uncertainty rather than returning an overconfident wrong answer, and invokes collaboration only in these flagged cases. In Example 1, the router's initial prediction (YES) is incorrect, but the system correctly flags its uncertainty and recruits specialist reasoning, which successfully corrects the error to the ground-truth answer (NO). Notably, MDAgent deploys dense collaboration (34 calls) yet still produces an incorrect final answer, while MOA and MedAgent achieve correctness at substantially higher computational cost (4--12 calls vs.\ \method's 4 calls, but with 1.4--6.1$\times$ more tokens). In Example 2, \method's initial uncertain prediction happens to be correct and is confirmed by collaboration with minimal overhead (4 calls, 38.34s), whereas MOA---despite also using 4 calls---produces an incorrect answer (D), and MDAgent and MedAgent require $3$--$4\times$ more calls and time to reach the correct conclusion. These cases illustrate that dense collaboration alone does not guarantee correctness; in contrast, uncertainty-aware routing targets collaboration precisely where it is most needed, improving both safety and efficiency by reducing overconfident errors while maintaining or improving diagnostic accuracy.

\newpage

\begin{figure}[!htb]
  \centering
  \hfill
  \begin{subfigure}[b]{0.3\textwidth}
    \centering
    \includegraphics[width=\linewidth]{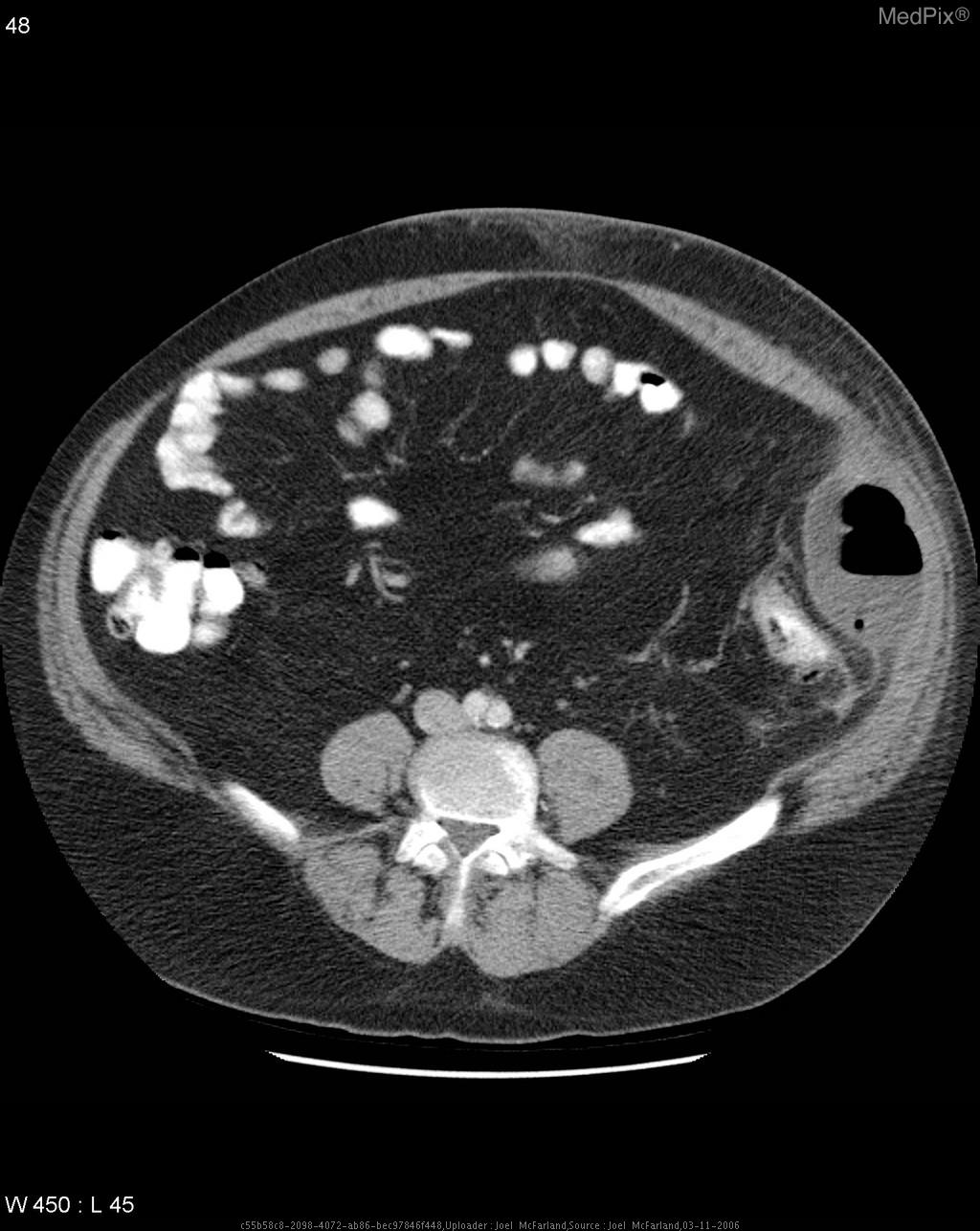}
    \caption{L1}
  \end{subfigure}
   \begin{subfigure}[b]{0.3\textwidth}
    \centering
    \includegraphics[width=\linewidth]{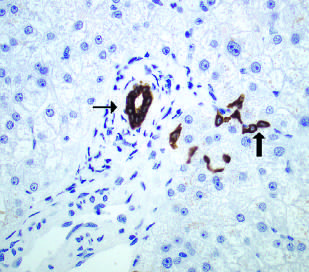}
    \caption{L2 and L6}
  \end{subfigure}
  \hfill
    \begin{subfigure}[b]{0.3\textwidth}
    \centering
    \includegraphics[width=\linewidth]{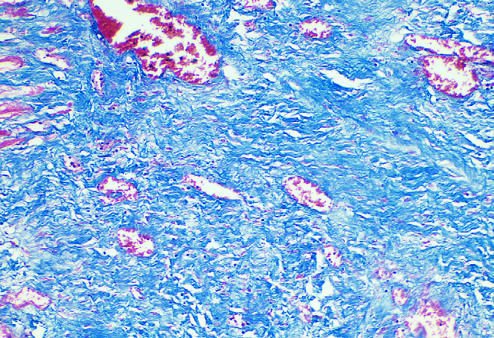}
    \caption{L3}
  \end{subfigure}
  \begin{subfigure}[b]{0.65\textwidth}
    \centering
    \includegraphics[width=\linewidth]{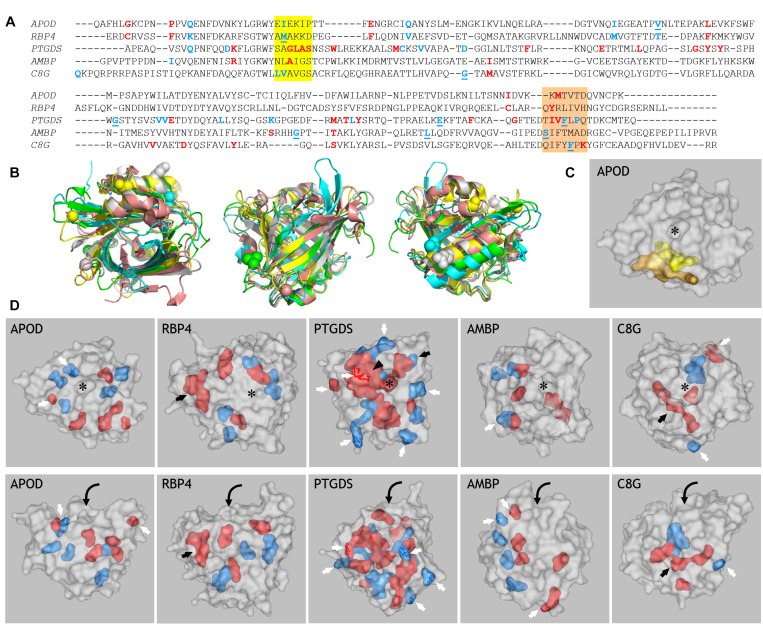}
    \caption{L4}
  \end{subfigure}
  \hfill
  \begin{subfigure}[b]{0.25\textwidth}
    \centering
    \includegraphics[width=\linewidth]{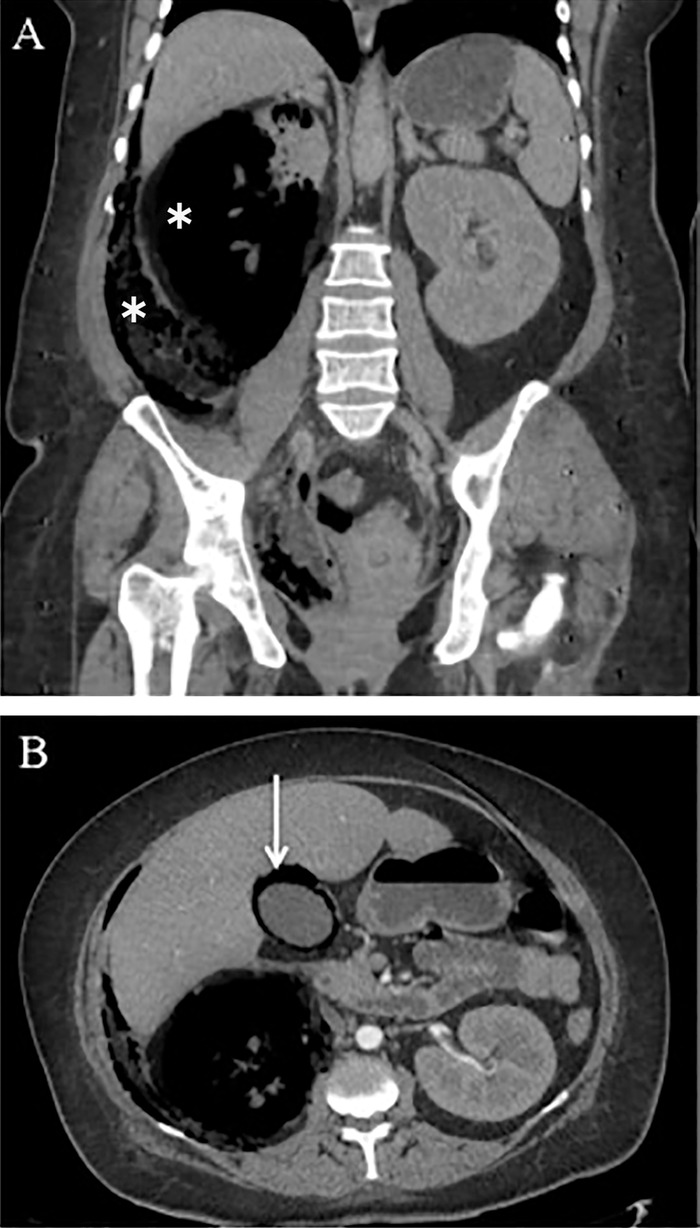}
    \caption{L5}
  \end{subfigure}
\caption{The medical images used in the Listing \ref{lst:example_1} to Listing \ref{lst:example_6}. Notably, the image (b) is used for two queries, a multi-choice question (L2) and a free-form question (L6).}
  \label{fig:examples}
\end{figure}

\section{Showcases of \dataset}
\label{sec:examples}
We present and analyze samples from \dataset to showcase the diversity, reasoning quality, and the reliability of multi-dimensional annotations. The corresponding images are shown in Fig.~\ref{fig:examples}.

The selected cases span multiple modalities and knowledge domains, including histopathology (Listings~\ref{lst:example_2}, \ref{lst:example_3}, and \ref{lst:example_6}), radiology (Listings~\ref{lst:example_1} and \ref{lst:example_5}), and structural biology (Listing~\ref{lst:example_4}), demonstrating the broad coverage of \dataset. Importantly, the dataset captures a wide spectrum of reasoning behaviors. For instance, Listing~\ref{lst:example_2} shows a case where the model reaches the correct conclusion (CK7 staining of biliary structures) but introduces a domain-specific factual error, reflecting realistic partial understanding. Listing~\ref{lst:example_1} illustrates a low-quality response with minimal reasoning and weak image grounding despite a correct final answer, while Listing~\ref{lst:example_3} provides a strong positive example with accurate domain knowledge and precise visual alignment in identifying collagen under trichrome staining. Listing~\ref{lst:example_4} further highlights cases where answers are correct but supported by weak or generic reasoning, whereas Listing~\ref{lst:example_5} demonstrates high-quality spatial reasoning and robust image grounding in a radiological context. Finally, Listing~\ref{lst:example_6} captures a subtle semantic discrepancy, where the model’s description (bile duct and ductules) is close but not strictly equivalent to the ground-truth concept (bile duct cells and canals of Hering), reflecting the fine-grained nature of medical terminology.

Meanwhile, for Listing~\ref{lst:example_5}, we also compare it with the doctor evaluation (\ie Listing~\ref{lst:example_7}). Both responses reach the exact same conclusion and share the same key reasoning: the gas is located outside the necrotic right kidney within the retroperitoneal space, while the gallbladder, left kidney, and stomach are all anatomically incorrect locations. The main differences lie in style and structure: the physician's response is more detailed and pedagogical, presenting a clear step-by-step elimination of each distractor without needing to cross-reference Fig.~\ref{fig:examples}(e)-B. Although the MLLM's response is more direct in reaching the answer, it still contains the key analytical reasoning chain (\eg step-by-step elimination), demonstrating that our constructed dataset is well-equipped with human-doctor-level reasoning chains.

For the associated quality annotations in \dataset, including logical consistency, factual correctness, and image alignment—closely match expert judgment across these varied scenarios. They successfully distinguish between correct answers with flawed reasoning (Listings~\ref{lst:example_2} and \ref{lst:example_4}), unsupported predictions (Listing~\ref{lst:example_1}), high-quality reasoning (Listings~\ref{lst:example_3} and \ref{lst:example_5}), and semantically imprecise outputs (Listing~\ref{lst:example_6}). This demonstrates that \dataset not only provides diverse and clinically relevant VQA samples, but also enables nuanced and reliable evaluation of model behavior. Overall, \dataset serves as a challenging and well-calibrated benchmark for advancing medical multimodal reasoning.

\begin{center}
\begin{minipage}{\textwidth}
\begin{lstlisting}[language=json,caption={Example 1},label={lst:example_1}]
  {
    "question": "Are air fluid levels depicted in this image?",
    "image-path": "synpic28718.jpg",
    "response": "I'm not 100% sure for my analysis to this medical image and the given question. But, I can guess the answer is YES. \n\n<answer>YES</answer>",
    "label": "YES",
    "logic_consistency": 1,
    "factual_correctness": 3,
    "image_alignment": 1,
    "difficulty": 2,
    "judgment": "The assistant provides essentially no reasoning and explicitly states uncertainty, then gives a bare YES answer, so the conclusion is unsupported and logically weak. Factual correctness is difficult to verify from the response alone; the image does appear to show intraluminal gas-fluid interfaces, so the final answer may be correct, but the lack of explanation and hedging prevent a higher score. Image alignment is poor because the assistant does not reference any specific visual features or regions in the CT image. The task itself is a basic image-identification question requiring limited medical knowledge, so difficulty is low."
  }
\end{lstlisting}
\end{minipage}
\end{center}

\newpage

\begin{center}
\begin{minipage}{\textwidth}
\begin{lstlisting}[language=json,caption={Example 2},label={lst:example_2}]
{
    "question": "Are bile duct cells and canals of Hering stained here with an immunohistochemical stain for cytokeratin 7?",
    "image-path": "train_0422.jpg",
    "response": "The image displays a histological section with cells stained to reveal specific structures. The presence of darkly stained, duct-like structures (indicated by arrows) amidst a background of smaller, round, blue-stained nuclei is consistent with the appearance of bile ducts and canals of Hering under immunohistochemical staining. Cytokeratin 7 (CK7) is a well-known marker that stains bile duct epithelium and canals of Hering, which are typically positive for CK7. The staining pattern in the image, with strong, specific labeling of the ductal structures, aligns with the expected immunohistochemical profile for CK7. The surrounding cells appear to be hepatocytes, which are also known to express CK7, but the most prominent staining is in the ductal structures. Given this evidence, the staining is consistent with an immunohistochemical stain for cytokeratin 7.<answer>YES</answer>"
    "label": "YES",
    "logic_consistency": 3,
    "factual_correctness": 2,
    "image_alignment": 4,
    "difficulty": 2,
    "judgment": "The response gives a coherent rationale linking brown immunostaining of duct-like structures to CK7-positive bile ducts/canals of Hering, so the logic is generally understandable but somewhat overstated and not fully rigorous. Image grounding is reasonably good because it refers to the dark brown ductal structures indicated by arrows against blue nuclei, which matches the visual features. However, factual accuracy is reduced by the claim that surrounding hepatocytes are also known to express CK7; in routine liver histology, hepatocytes are generally CK7-negative, while bile duct epithelium and canals of Hering are CK7-positive. That is a significant pathology-specific error, though the final YES answer is likely correct. The question itself is a basic histopathology/IHC identification task requiring limited specialist knowledge, so difficulty is low-to-moderate."
}
\end{lstlisting}
\end{minipage}
\end{center}

\begin{center}
\begin{minipage}{\textwidth}
\begin{lstlisting}[language=json,caption={Example 3},label={lst:example_3}]
  {
    "question": "Is Collagen stained blue by the trichrome stain?",
    "image-path": "train_0473.png",
    "response": "In histological staining, the trichrome stain is commonly used to differentiate collagen from other tissue components. In this stain, collagen fibers are typically stained blue or green, while muscle fibers appear red or pink, and nuclei are often counterstained with a different color (e.g., purple or black).\n\nIn the provided image, the background matrix appears predominantly blue, and there are scattered pinkish-purple structures that resemble nuclei or other cellular components. The blue-stained areas have a fibrous, wavy appearance consistent with collagen fibers.\n\nTherefore, based on the visual evidence and standard histological knowledge, collagen is stained blue by the trichrome stain.\n\n<answer>YES</answer>",
    "label": "YES",
    "logic_consistency": 5,
    "factual_correctness": 5,
    "image_alignment": 5,
    "difficulty": 3,
    "judgment": "The assistant provides a clear, step-by-step explanation linking trichrome staining principles to the image features, correctly identifying blue-stained fibrous areas as collagen. The description of other components (pink nuclei, background) aligns with standard histology. The response is factually accurate and demonstrates precise visual analysis of the image, though the question itself is moderate in complexity, requiring basic histological knowledge and single-step reasoning."
  }
\end{lstlisting}
\end{minipage}
\end{center}

\begin{center}
\begin{minipage}{\textwidth}
\begin{lstlisting}[language=json,caption={Example 4},label={lst:example_4}]
  {
    "question": "What do the colored spheres in Figure B represent? A: Ligand pockets. B: Beta-barrels. C: Type-I residues. D: SDP residues.",
    "image-path": "PMC8420045_F3.jpg",
    "response": "I can discern various colored spheres across the protein structures shown in Figure B. These spheres seem to highlight specific types of residues based on their chemical characteristics.\n\nFrom the image:\n- The spheres are shown in different colors (such as red, blue, and white), which could indicate specific features of the residues.\n- The context of the surrounding figures (which display proteins and their structural features) suggests that the colored spheres are likely highlighting specific residues that are crucial for particular interactions or functions.\n\nGiven the evidence in the image and the provided options, the most plausible conclusion is that these spheres represent **Type-I residues**, as they are typically associated with specific amino acid positions critical for structural functions in protein analysis.\n\n<think>Based on the visualization and the context of protein analysis, the colored spheres in Figure B most likely represent Type-I residues.</think><answer>C</answer>",
    "label": "C",
    "logic_consistency": 5,
    "factual_correctness": 5,
    "image_alignment": 4,
    "difficulty": 3,
    "judgment": "The response correctly identifies the answer as C (Type-I residues), matching the label, so factual correctness is full. The internal reasoning is coherent and non-contradictory, though it lacks strong evidence from the figure; logic consistency is still high. Without seeing the image, alignment is judged as reasonable (4) because the description of colored spheres is plausible for a protein figure, and the correct choice suggests good visual matching. Difficulty is moderate as the question requires knowledge of protein structural features but is not extremely specialized."
  }
\end{lstlisting}
\end{minipage}
\end{center}

\begin{center}
\begin{minipage}{\textwidth}
\begin{lstlisting}[language=json,caption={Example 5},label={lst:example_5}]
  {
    "question": "What is the bottom asterisk pointing to in the image A? A:Intramural gas in the gallbladder B: Gas extending through the retroperitoneum C: Extensive parenchymal destruction of the left kidney D: Inflammation in the stomach.",
    "image-path": "PMC8442220_f2.jpg",
    "response": "<think>The coronal CT image A shows the patient's right side on the left of the image. There is a markedly abnormal right renal area with extensive gas replacing much of the renal parenchyma. The upper asterisk is within this gas-filled destructive renal region. The bottom asterisk points to streaky/mottled gas tracking outside the kidney along the right flank/retroperitoneal space. This is not the gallbladder (seen in image B with arrow), not the left kidney, and not the stomach.</think><answer>B</answer>",
    "label": "B",
    "logic_consistency": 4,
    "factual_correctness": 4,
    "image_alignment": 5,
    "difficulty": 3,
    "judgment": "The assistant gives a clear rationale: it distinguishes the lower asterisk from the upper renal gas collection, notes that the lower marker indicates gas tracking outside the kidney along the flank/retroperitoneum, and explicitly rules out the gallbladder, left kidney, and stomach. This is logically coherent and matches the visual appearance of image A. The answer choice B is medically appropriate; the image shows emphysematous change with gas extending into the retroperitoneal/perinephric region rather than gallbladder gas, left renal destruction, or gastric inflammation. Image grounding is strong because the response localizes the finding to the specific lower asterisk and relates it to surrounding anatomy. A perfect logic/factual score is not warranted because the reasoning is concise rather than exhaustive, and it references image B to exclude the gallbladder even though the question asks only about image A. The question itself is of moderate difficulty, requiring standard radiologic anatomy and pathology recognition rather than highly advanced inference."
  }

\end{lstlisting}
\end{minipage}
\end{center}

\begin{center}
\begin{minipage}{\textwidth}
\begin{lstlisting}[language=json,caption={Example 6},label={lst:example_6}]
  {
    "question": "What are stained here with an immunohistochemical stain for cytokeratin 7?",
    "image-path": "train_0422.jpg",
    "response": "<think>The image is a micrograph of liver tissue where an immunohistochemical stain (brown) has been applied, likely for cytokeratin 7. The thin arrow points to a distinct, lumen-containing structure within a portal tract, which is morphologically characteristic of a bile duct. The thick arrow points to smaller, irregular clusters of stained cells at the periphery of the portal tract, representing bile ductules. Both of these biliary epithelial structures are known to be positive for cytokeratin 7, while the surrounding hepatocytes are negative.</think><answer>A bile duct (thin arrow) and bile ductules (thick arrow)</answer>",
    "label": "BILE DUCT CELLS AND CANALS OF HERING",
    "logic_consistency": 5,
    "factual_correctness": 3,
    "image_alignment": 3,
    "difficulty": 3,
    "judgment": "The response is internally logical and clearly describes the two arrow labeled structures. However, the label specifies 'bile duct cells and canals of Heringm' while the answer says 'bile duct and bile ductules', bile ductules are not exactly the same as canals of Hering (the latter are transitional channels between canaliculi and ductules), so factual correctness is slightly reduced. Without seeing the actual image, the alignment score is a moderate 3, assuming the arrows might not perfectly match the labeled structures. Difficulty is average (3) because CK7 staining in liver is a standard histopathology concept."
  }

\end{lstlisting}
\end{minipage}
\end{center}

\begin{center}
\begin{minipage}{\textwidth}
\begin{lstlisting}[language=json,caption={Human Doctor Evluation to Example 5},label={lst:example_7}]
  {
    "question": "What is the bottom asterisk pointing to in the image A? A:Intramural gas in the gallbladder B: Gas extending through the retroperitoneum C: Extensive parenchymal destruction of the left kidney D: Inflammation in the stomach.",
    "image-path": "PMC8442220_f2.jpg",
    "doctor": "(1) Image Localization: Figure A is a coronal CT image, with the left side of the image corresponding to the patient's right side. A large area of air-containing low-attenuation shadow can be seen in the right renal region, indicating renal parenchymal necrosis with gas accumulation. The upper asterisk marks the intrarenal gas and necrosis, while the lower asterisk marks the extrarenal gas shadow located in the retroperitoneal space.

    (2) Systematic elimination of distractors!
    A (Intramural gas in the gallbladder wall): The gallbladder is located in the right upper quadrant liver bed area, which does not match the anatomical location of the lower asterisk. The arrow in Figure B merely serves as a reference for the gallbladder and can be excluded without cross-referencing between images.
    C (Extensive parenchymal destruction of the left kidney): The left kidney is located on the right side of the image, with essentially normal morphology and density, showing no gas accumulation or extensive destruction. The lesion is entirely on the right side and can therefore be excluded.
    D (Gastric inflammation): The main body of the stomach is in the left upper quadrant, posterior to the liver; the gastric antrum is near the right upper quadrant, adjacent to the pancreatic head. This does not correspond to the right-sided retroperitoneal location of the lower asterisk and can be excluded.
    
    (3) Conclusion: The gas indicated by the lower asterisk extends outward from the necrotic right kidney into the retroperitoneal space, corresponding to option B.",
  }

\end{lstlisting}
\end{minipage}
\end{center}

\section{Prompts Design}
\label{sec:prompts}
\paragraph{Prompts for Building \dataset.} \dataset is a reasoning-centric dataset, and the reasoning quality is critical. To ensure the quality, we employ an independent judge MLLM (\ie GPT 5.4) to evaluate the quality of each reasoning chain. We design the prompts for the judge model as below:
\begin{tcolorbox}[colback=cadmiumgreen!5!white,colframe=forestgreen!75!white,breakable,title=Prompts for the judge MLLM]
\begin{VerbatimWrap}
You are an expert evaluator for multimodal medical QA tasks. Your role is to critically assess the assistant's response to the provided User Input.

### User's Question 
{question}

### Assistant Response 
{assistant_response}

---

## Evaluation Rubrics

Evaluate the assistant's response on the following four dimensions. Each dimension is scored on a 1-5 scale.

### 1. Logic Consistency (1-5)
Assesses the coherence, structure, and validity of the reasoning chain.

| Score | Criteria |
|-------|----------|
| 1 | Severe logical fallacies, self-contradiction, circular reasoning, or complete lack of reasoning |
| 2 | Flawed reasoning with significant gaps; conclusion does not logically follow from premises |
| 3 | Generally sound reasoning but lacks depth, contains minor leaps, or oversimplifies complexity |
| 4 | Clear, structured reasoning that follows medical diagnostic/analytical standards |
| 5 | Flawless, step-by-step deductive reasoning; robust, complete, and clinically sound |

### 2. Factual Correctness (1-5)
Assesses the medical accuracy of the response.

| Score | Criteria |
|-------|----------|
| 1 | Critical hallucinations; dangerous or completely incorrect medical advice/claims |
| 2 | Significant factual errors regarding symptoms, anatomy, pathology, or treatment |
| 3 | Mostly correct but contains minor inaccuracies, outdated information, or imprecise terminology |
| 4 | Factually correct throughout; no errors detected |
| 5 | Factually correct and precise; uses professional terminology accurately and appropriately |

### 3. Image Alignment (1-5)
Assesses how well the response utilizes and accurately interprets visual information.

| Score | Criteria |
|-------|----------|
| 1 | Completely ignores the image; describes features not present or contradicts visual evidence |
| 2 | Mentions the image but misinterprets key visual features (e.g., wrong location, wrong modality, incorrect identification) |
| 3 | Generic response that could apply to any similar image; lacks specific visual grounding |
| 4 | Correctly identifies and describes relevant visual features; demonstrates basic visual grounding |
| 5 | Deep visual grounding; precisely correlates specific image regions with clinical reasoning; demonstrates expert visual analysis |

### 4. Difficulty (1-5)
Assesses the complexity of the question itself (not the answer quality).

| Score | Criteria |
|-------|----------|
| 1 | Simple fact retrieval or common knowledge; no medical expertise required |
| 2 | Basic identification task; requires minimal medical knowledge |
| 3 | Moderate complexity; requires standard medical knowledge and single-step reasoning |
| 4 | Complex case; requires multi-step reasoning, integration of multiple concepts, or specialist knowledge |
| 5 | Highly complex; requires integrating multiple modalities, rare/specialized knowledge, or advanced differential diagnosis |

---

## Strict Evaluation Guidelines

1. **Be conservative**: A score of 5 is reserved for "perfect" or "exemplary" responses. Use 4 for "good but not perfect."

2. **Penalize hallucinations heavily**: Any fabricated information should result in a Factual Correctness score smaller than 2.

3. **Guess vs. reasoning**: If the reasoning is missing, flawed, or insufficient, but the final answer is correct (guessed), cap the Logic Consistency score at 2.

4. **Image requirement**: A response that fails to meaningfully engage with the image (when visual input is essential) cannot receive an Image Alignment score above 3.

5. **Response completeness**: If the response is incomplete or refuses to answer without justification, scores should reflect this deficiency.

6. **Evidence-based judgment**: All scores must be justified with specific evidence from the assistant's response.

---

## Output Format

Output ONLY a valid JSON object. Do not include markdown formatting, code blocks, or explanations outside the JSON.

{{
    "scores": {{
        "logic_consistency": <int 1-5>,
        "factual_correctness": <int 1-5>,
        "image_alignment": <int 1-5>,
        "difficulty": <int 1-5>
    }},
    "critique": "<brief, evidence-based justification for your scores>"
}}

\end{VerbatimWrap}
\end{tcolorbox}

\paragraph{Prompts for Building \method.} In our method, there are three kinds of roles, undertaking different tasks: (1) Router: the agent for initial assessment, certainty check and expert recruitment; (2) Specialist: the agent is assigned with specific tasks and collaboration patterns in the multi-agent team; (3) Moderator: the agent aims to summarize the analysis reports from multiple agents. Their system prompts, role prompts and task prompts are designed as follows:
\begin{tcolorbox}[colback=cadmiumgreen!5!white,colframe=forestgreen!75!white,breakable,title=Prompts for the router agent]
\begin{VerbatimWrap}
# System Prompt
You are a medical resource allocation router. Your task is to allocate the medical question to the most appropriate specialist agents efficiently and effectively.

## Task
Please recruit medical experts for this question: <image> {question}

## Instruction
1. Recruit maximally 5 specialist roles relevant to answer this question. Reference roles: [Radiologist, Pathologist, Surgeon, Oncologist, Endocrinologist, Geneticist, Cardiologist, Neurologist]
2. Generate a brief task description for each role explaining their job in the collaboration to correctly answer this question.
3. For role descriptions, using **|** to seperate different roles.

## Output Format
<role>[Role1, Role2, ...]</role>\n<role_desc>[Role1: Description | Role2: Description | ...]</role_desc>

## Example
<role>Radiologist, Oncologist</role>\n<role_desc>Interprets CT images to detect lung masses | Evaluates malignancy and treatment strategy</role_desc>
\end{VerbatimWrap}
\end{tcolorbox}

\begin{tcolorbox}[colback=cadmiumgreen!5!white,colframe=forestgreen!75!white,breakable,title=Prompts for the specialist agent]
\begin{VerbatimWrap}

# System Prompt
Your task is {role_description}, and your job is {task_description}.

## Task
Analyze the provided medical image and answer the following {answer_type} question: <image> 
{question}
## Instructions
1. **Visual Analysis**: Thoroughly examine the image and identify all anatomical structures, abnormalities, or features relevant to the question.
2. **Evidence-Based Reasoning**: Base your conclusion strictly on the visual evidence. Cite specific observations that support your analysis.
3. **Uncertainty Annotation**: If the evidence is genuinely ambiguous or critical features are obscured, return UNCERTAIN in your response.
## Output Format
<reason>[Your visual analysis citing specific evidence]</reason><certainty>[CERTAIN/UNCERTAIN]</certainty><answer>[Your definitive best-effort answer: {answer_type}]</answer>
·
## Examples

**High Confidence Case:**
<reason>The CT slice clearly shows a hyperdense, well-circumscribed lesion in the right lung base with characteristics typical of a hamartoma.</reason><certainty>CERTAIN</certainty><answer>Benign</answer>

**Low Confidence Case:**
<reason>The image shows a possible mass in the mediastinum, but the slice level appears to be above the cardiac silhouette and the structure boundaries are poorly defined due to motion artifact.</reason><certainty>UNCERTAIN</certainty><answer>Malignant</answer>

## Critical Rules
1. The `<certainty>` tag MUST be either `CERTAIN` or `UNCERTAIN` (no other values allowed).
2. The `<answer>` tag MUST always contain a specific {answer_type} value, regardless of certainty level.
\end{VerbatimWrap}
\end{tcolorbox}

\begin{tcolorbox}[colback=cadmiumgreen!5!white,colframe=forestgreen!75!white,breakable,title=Prompts for the moderator agent]
\begin{VerbatimWrap}
# System Prompt
You are a medical assistant who excels at summarizing and synthesizing based on multiple experts from various domain experts.

## Context
**Image**: <image>
**Question**: {question}
**Answer Type**: {answer_type} (Your final answer MUST strictly conform to this type)

**Initial Assessment**:
{initial_assessment}

**Expert Reports**:
{report}

## Analysis Framework

### 1. Individual Assessment
For the given initial assessment and expert reports, evaluate:
- **Certainty Level**: CERTAIN or UNCERTAIN (from <certainty> tag)
- **Key Evidence**: Specific visual features cited from the image
- **Reasoning Quality**: Logical soundness of the inference chain
- **Risk of Hallucination**: Whether claims exceed visible evidence

### 2. Cross-Expert Comparison
- **Consensus Detection**: Identify agreements and disagreements across experts and the initial assessment
- **Evidence Strength**: Which conclusion has stronger visual support?
- **Certainty Weighting**: Prioritize CERTAIN answers unless their evidence is weaker
- **Conflict Analysis**: If experts disagree, determine if the disagreement stems from:
    - Different interpretation of the same evidence → weigh reasoning quality
    - Different evidence focus → check which evidence is more relevant to the question
    - Fundamental ambiguity in the image → acknowledge irreducible uncertainty

### 3. Adjudication Rules (Apply in Order)
1. **Strong Consensus**: If there are more than two experts certainly agree with strong evidence, adopt their answer
2. **Quality Override**: If UNCERTAIN experts have stronger evidence than CERTAIN ones → favor evidence over certainty label

### 4. Meta-Evaluation (Critical Step)
Even if experts agree with the initial assessment, ask yourself:
- Is there a **plausible alternative interpretation** they missed?
- Does their conclusion **overreach the visual evidence**?
- Would a **different expert specialty** (e.g., radiologist vs. pathologist) disagree?
Finally, your task is to determine the answer of this question according to your analysis to the initial assessment and expert reports.

## Output Requirements
**CRITICAL**: Your answer MUST be a valid {answer_type} value.

## Output Format
<reason>[Step-by-step adjudication: summarize expert positions, compare evidence quality, explain your reasoning for selecting the final answer]</reason><answer>[Your final answer]</answer>

\end{VerbatimWrap}
\end{tcolorbox}

\section{Limitation and Social Impact}

\paragraph{Limitation.}
In \method, we employ local small-scale MLLMs as the agent. Although it can preserve the privacy, save the computational cost and latency, the overall performance is still bounded by the capacity of the underlying backbone model, which may limit its ability to handle extremely complex or rare medical cases compared to very large proprietary models. Therefore, how to deal with extremely complex or rare medical cases can be a potential future direction. 

\paragraph{Social Impact.}
This work has important implications for the development of privacy-preserving and deployable medical AI systems. By enabling fully local inference and reducing reliance on cloud-based models, \method helps mitigate data privacy and regulatory concerns in clinical environments. Moreover, the integration of uncertainty-aware decision-making can improve the reliability and safety of AI-assisted diagnosis by reducing overconfident errors. However, potential risks remain: incorrect or biased reasoning generated by the model could still mislead practitioners if used without proper oversight, and the dataset construction process may inherit biases from source data and teacher models. Therefore, \method should be deployed as a decision-support tool rather than a replacement for medical professionals, and future work should focus on improving robustness, fairness, and clinical validation.



\end{document}